\documentclass{article}

\PassOptionsToPackage{numbers,compress}{natbib}
\usepackage[preprint]{neurips_2026}
\usepackage{wrapfig}
\usepackage{graphicx} 
\usepackage{graphicx}
\usepackage{amsmath}
\usepackage{amssymb}
\usepackage{amsthm}
\usepackage{multirow}
\usepackage{array}
\usepackage{float}
\usepackage{caption}
\usepackage[utf8]{inputenc}
\usepackage[T1]{fontenc}
\usepackage{hyperref}
\usepackage{url}
\usepackage{booktabs}
\usepackage{amsfonts}
\usepackage{nicefrac}
\usepackage{microtype}
\usepackage{xcolor}
\usepackage{colortbl}
\usepackage{algorithm}
\usepackage[noend]{algpseudocode}

\usepackage[capitalize]{cleveref}
\crefname{section}{Sec.}{Secs.}
\Crefname{section}{Section}{Sections}
\Crefname{table}{Table}{Tables}
\crefname{table}{Tab.}{Tabs.}
\crefname{algorithm}{Alg.}{Algs.}
\Crefname{algorithm}{Algorithm}{Algorithms}

\theoremstyle{plain}

\theoremstyle{definition}

\newcommand{\R}{\mathbb{R}}

\title{Efficient Test-Time Finetuning of LLMs via Convex Reconstruction and Gradient Caching}
\author{
  Alaa Khamis \\
  Department of Computer Science \\
  University of Haifa \\
  \And
  Alaa Maalouf \\
  Department of Computer Science \\
  University of Haifa
}

\begin{document}

\maketitle

\begin{abstract}
Test-time finetuning (TTFT) is a rapidly evolving paradigm that adapts a language model to each prompt by retrieving related sequences, updating the model on them, and then evaluating the prompt.
However, TTFT is only practical if it is \emph{fast}: selection and finetuning both happen per query, making each a direct bottleneck. Existing methods trade speed for quality: fast retrieval is often redundant, while stronger diversity-aware selection adds prohibitive per-query cost.
We introduce \textbf{HullFT}\footnote{Code available at \url{https://github.com/alaa-khamis/HullFT}.}, a geometric approach to TTFT that addresses both bottlenecks. Given a query, HullFT first represents the query embedding as a sparse convex combination of few training sequences, using efficient projection-free Frank--Wolfe optimization. This yields a support set that is inherently \emph{relevant and diverse}. We then convert the fractional convex weights into an exact integer multiset for finetuning through a geometric integerization procedure. The resulting multiplicities naturally create repeated examples, which we exploit with \textbf{Gradient Reuse} to amortize forward--backward computation across repeated finetuning steps. Our experiments show that HullFT improves the quality--efficiency tradeoff over current state-of-the-art TTFT methods, achieving lower bits-per-byte at substantially lower total runtime.
\end{abstract}

\section{Introduction}
\label{sec:intro}

Large language models are trained on web-scale corpora and optimized for broad,
general-purpose performance across a vast distribution of text~\cite{raffel2020t5, penedo2023refinedweb}.
As a result, their weights represent a global optimum that is unlikely to be
locally optimal for any specific input.
Test-time training (TTT)~\cite{sun2020ttt, gandelsman2022ttt_mae, hardt2024ttt_nn, sun2025ttt_rnn, akyurek2024ttt_fewshot}, also referred to as test-time finetuning (TTFT) in recent influential work~\cite{hubotter2025sift}, addresses this directly:
at inference time, given a prompt $q$, we retrieve the $N$ most relevant sequences from a large corpus, take one gradient step per retrieved sequence, and evaluate the finetuned model on $q$.
Retrieving as few as 20 neighbors is enough to substantially close the gap between models differing by more than an order of magnitude in parameter count~\cite{hardt2024ttt_nn}.

Data selection for TTFT runs per prompt at inference time, making it a direct contributor to user-facing latency.
Current methods sit at opposite ends of a quality--efficiency tradeoff. Pure nearest-neighbor retrieval~\cite{hardt2024ttt_nn} returns the top-$N$ candidates (nearest neighbors in some embedding space) via a FAISS index~\cite{johnson2021faiss}: fast, but blind to redundancy. 
On large corpora, duplicate content is common~\cite{wenzek2020ccnet, abbas2023semdedup}; without accounting for redundancy, the top-$N$ neighbors can collapse to near-identical sequences, causing every subsequent gradient step to repeat the same signal. Diversity-aware selection methods address this by explicitly penalizing redundancy when choosing examples~\cite{hubotter2025sift}, producing more informative batches. The improvement in bits-per-byte (bpb)~\cite{gao2020pile} over pure retrieval is significant, though such greedy selection introduces overhead, especially at lower $N$ values.

\textbf{Novelty. }In this work, we take a different route, drawing on provable sparse convex approximation.
The key idea is that directions in the embedding space carry meaning:
sequences spanning different directions cover a broader range of semantic features, while sequences pointing in nearly the same direction are largely redundant~\cite{carbonell1998mmr, kulesza2012dpp}. Thus, framing data selection as a \emph{convex approximation} of the prompt $q$ means diversity is inherently derived from the geometry, without explicit penalty terms, greedy heuristics, or expensive search and heavy gradient descent based optimization methods.
Once a candidate is selected to help reconstruct $q$, near-duplicates pointing in the same direction are naturally down-weighted, as they offer little additional value.
The optimization naturally (inherently from the convex approximation definition) turns to unexplored directions instead, covering the prompt's context (in the embedding space) more broadly~\cite{sener2018active_coreset, mirzasoleiman2020coresets}.


Approximating $q$ as a convex combination of candidates is precisely the \emph{approximate Carath\'{e}odory problem}~\cite{barman2015approx_caratheodory, caratheodory1907variabilitatsbereich, combettes2023caratheodory}: find a small support set with positive weights summing to $1$ whose weighted centroid is close to $q$.
The approximate Carath\'{e}odory theorem guarantees that an $\varepsilon$-accurate solution (in squared $\ell_2$ error) using $O(1/\varepsilon)$ points always exists, regardless of ambient dimension~\cite{barman2015approx_caratheodory}.
This gives us a principled way to obtain a sparse, relevance-aware, and diversity-aware representation of the query from a kNN candidate pool. 
Notably, convex reconstruction alone does not yet specify a TTFT update sequence. It returns fractional weights over selected examples, while TTFT requires an explicit set of $N$ samples.
We therefore convert the weighted solution into integer multiplicities that preserve the geometric approximation as closely as possible, see more details in \cref{sec:app_abl_fw_family}.

\begin{figure}[t]
    \centering
    \includegraphics[width=.95\linewidth]{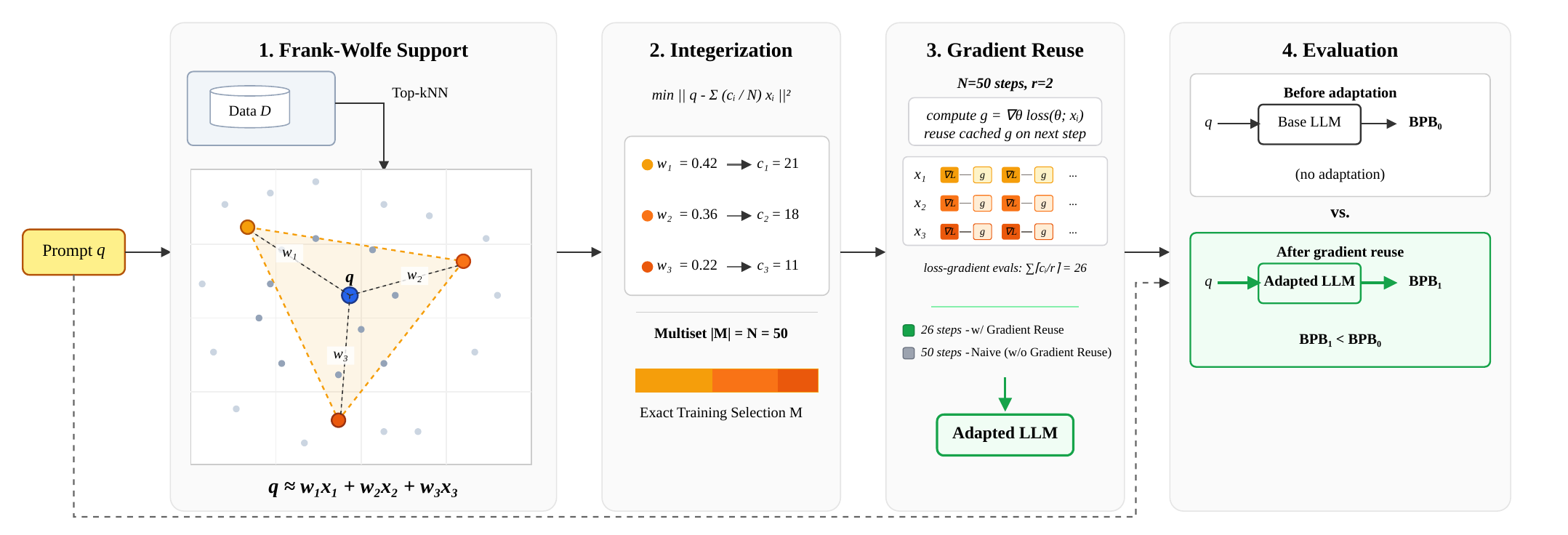}
    \caption{
        Our test-time finetuning pipeline. \textbf{1. Frank-Wolfe Support:} 
        Given a prompt $q$, we retrieve a candidate pool from the corpus, then approximate $q$ as a sparse convex combination to select a support set. \textbf{2.~Integerization:}     Fractional weights are converted to integer counts forming an exact $N$-point multiset. \textbf{3.~Finetuning \& Inference:} The base LLM is finetuned on this multiset before evaluating $q$ via gradient reusing on repeated samples. 
    }
    \label{fig:pipeline}
\end{figure}


Crucially, this discrete reconstruction does more than make the method executable.
Once the fractional weights are converted into integers, some selected examples appear multiple times in the final multiset.
These repetitions create an opportunity for efficient finetuning: rather than recomputing a fresh forward--backward pass for every repeated occurrence, we can reuse cached gradients across short runs of identical examples.
Thus, integerization not only bridges geometry and TTFT, but also induces the structure that enables Gradient Reuse. \Cref{fig:pipeline} summarizes the HullFT pipeline. 

\textbf{Our contributions.}
We introduce \textbf{HullFT}, an efficient test-time finetuning method that improves the quality--efficiency tradeoff of TTFT through three coupled ideas.
\begin{enumerate}
     \item We formulate query/prompts-specific data selection as a sparse convex approximation problem over a kNN-retrieved candidate pool, yielding a support set that is both relevant and diverse. We suggest solving it using the \textbf{Frank--Wolfe} algorithm. This efficient projection-free convex approximation procedure selects a small, diverse, and relevant support set from the candidate pool (\cref{alg:fw} in the appendix).
    \item We introduce a \textbf{geometric integerization} procedure (\cref{alg:integerize}) that turns fractional convex weights into an exact $N$-example training multiset while preserving approximation quality. \label{cont:second}
    \item We show that the resulting multiset (from~\ref{cont:second})  enables \textbf{Gradient Reuse}, which amortizes gradient computation across repeated training examples and reduces finetuning cost (\cref{alg:grad_reuse}). 
\end{enumerate}

Across 12 subsets of \textbf{The Pile}~\cite{gao2020pile}, HullFT strictly outperforms our competing methods (kNN~\cite{hardt2024ttt_nn} and SIFT~\cite{hubotter2025sift}) on quality--efficiency tradeoffs, with the largest gains in the latency-critical regime.

\section{Related work}
\label{sec:related}

\textbf{Test-time training and finetuning.}
Instance-specific adaptation traces back to transductive learning~\cite{joachims1999transductive} and has recently re-emerged as test-time training (TTT), where deep networks adapt at inference time using self-supervised objectives~\cite{sun2020ttt, gandelsman2022ttt_mae, sun2025ttt_rnn, akyurek2024ttt_fewshot}. 
For language models, nearest-neighbor TTT finetunes on retrieved examples~\cite{hardt2024ttt_nn}, while SIFT adds an information-theoretic selection criterion that accounts for redundancy~\cite{hubotter2025sift}. 
Alternatively, retrieval-augmented models place retrieved text in context~\cite{lewis2020rag, borgeaud2022retro}, but self-attention cost limits context size and full benefits often require retrieval-aware training~\cite{hardt2024ttt_nn}. 
Related selection methods balance relevance and diversity via maximal marginal relevance or determinantal point processes~\cite{carbonell1998mmr, kulesza2012dpp}. 
Influence-based methods estimate example utility~\cite{hampel1974influence, koh2017understanding}, including through gradient similarity for LLM finetuning~\cite{pruthi2020tracin, kwon2024datainf, xia2024less}, but typically score points independently and thus miss redundancy~\cite{hubotter2025sift}. 
Active learning selects diverse subsets to cover the data manifold~\cite{sener2018active_coreset, mirzasoleiman2020coresets}, but is not query-conditioned.

\textbf{Data selection.}
Early transfer-learning work showed that an additional finetuning stage on in-domain text produces large gains across tasks~\cite{howard2018ulmfit, gururangan2020dontstop}.
The approach has since been validated in biomedicine~\cite{lee2020biobert, alsentzer2019clinicalbert, huang2019clinicalbert}, scientific text~\cite{beltagy2019scibert}, legal language~\cite{chalkidis2020legalbert}, and clinical question answering~\cite{singhal2023medpalm}, establishing domain-specific adaptation as a default recipe.
Beyond the choice of domain, the composition of the training subset itself can be decisive: a handful of curated examples sometimes matches full-corpus finetuning~\cite{zhou2023lima}, parameter-efficient methods make per-task adaptation practical even for very large models~\cite{hu2022lora}, and recent work explicitly optimizes which data to include for specialization~\cite{hulkund2025datas3, sreeram2025compress}.
Adaptation benefits extend beyond language; for instance, they have been shown to govern out-of-distribution robustness in vision-based driving~\cite{mallak2026robustness, wang2024drive}.
At the pre-training scale, a parallel effort in data curation tackles redundancy through deduplication, quality filtering, and domain reweighting~\cite{wenzek2020ccnet, abbas2023semdedup, sachdeva2024ask_llm, wettig2024qurating, raffel2020t5, penedo2023refinedweb}.
Our approach pushes this logic to its extreme: rather than curating a training set once, we curate query-specific subsets at inference time, using convex geometry to jointly enforce relevance and diversity within a single optimization.

\textbf{Convex approximation, Frank--Wolfe, and coresets.}
Carath\'{e}odory's theorem~\cite{caratheodory1907variabilitatsbereich} guarantees that any point in the convex hull of a set in $\R^d$ can be written as a convex combination of at most $d{+}1$ points, requiring $O(n^2d)$ time for constructing such a set. Lately, ~\cite{maalouf2019fast_lms} proposed a fast construction running in $O(nd + d^4 \log n)$ time. 
The approximate variant relaxes this to $O(1/\varepsilon^2)$ points independent of dimension~\cite{maurey1972lemma, barman2015approx_caratheodory, mirrokni2017tight_caratheodory}.
The Frank--Wolfe (conditional gradient) algorithm~\cite{frank1956algorithm, jaggi2013revisiting} provides a constructive proof of this result, with each iteration being a sparse convex combination that grows by at most one point~\cite{combettes2023caratheodory}. Other variants achieve faster convergence under favorable geometry~\cite{guelat1986fw_linear, lacoste2016fw_survey}.
The connection between Frank--Wolfe and coresets (small weighted subsets that provably approximate a cost function over the full data) was established by Clarkson~\cite{clarkson2010coresets_fw} and developed in a rich body of work on geometric data summarization~\cite{agarwal2005coresets, har2011geometric, phillips2017coresets}.
In machine learning, coreset methods have been applied to regression~\cite{maalouf2019fast_lms, maalouf2021coresets_finite}, data-efficient model training and subset selection for deep learning~\cite{mirzasoleiman2020coresets, sener2018active_coreset, tukan2023provable, killamsetty2021glister, killamsetty2021gradmatch, coleman2020selection}, model compression and network pruning~\cite{tukan2022pruning, liebenwein2020provable, mussay2022data, baykal2022sipping}, and general-purpose loss approximation under unified coreset frameworks~\cite{maalouf2023autocoreset, maalouf2024unified, feldman2011unified, braverman2016new, bachem2017practical, munteanu2018coresets}.
Our selection stage instantiates Frank--Wolfe on the $\ell_2$-squared objective over the probability simplex, yielding a coreset of the candidate pool. Integerization then converts into a multiset for finetuning.

\section{Method}
\label{sec:method}


\textbf{Notation.}
Let $\langle u, v\rangle = u^\top v$ be the dot product on $\R^d$, $\mathbf{1}[\cdot]$ the indicator function ($\mathbf{1}[A]=1$ if predicate $A$ is true, $0$ otherwise), $\Delta^K$ the probability simplex in $\R^K$, and $e_i\in\R^K$ the $i$-th standard basis vector.
We use $q\in\R^d$ to denote a query, $\{p_1,\ldots,p_K\}\subset\R^d$ to denote a candidate training pool, and $P\in\R^{d\times K}$ to denote the matrix whose $i$-th column is $p_i$, so $Pw = \sum_i w_i\,p_i$ for any $w\in\R^K$ (all in the embedding space).
The final selected support set is $S = \{s_1,\ldots,s_{|S|}\} \subseteq \{p_1,\ldots,p_K\}$, with associated integer counts $c \in \mathbb{Z}_{\geq 0}^{|S|}$, where $c_j$ is the count of $s_j$.
Throughout, $N$ is the finetuning budget (total multiset size), $\varepsilon$ is the Frank--Wolfe tolerance, and $m$ is the support cap. 
The base model has parameters $\theta$, learning rate $\eta$, and per-sequence loss $\mathcal{L}(\theta; s)$.

\subsection{Data selection via convex approximation}

\begin{algorithm}[t]
\caption{\textsc{Integerize}$(q, \{p_1,\ldots,p_K\}, w, N, T)$}
\label{alg:integerize}
\begin{algorithmic}[1]
\Statex \textbf{Input:} query $q$, candidate pool $\{p_1,\ldots,p_K\}$, sparse weights $w \in \Delta^K$, budget $N$, swap passes $T$ (default $2$)
\Statex \textbf{Output:} support set $S = \{s_1,\ldots,s_{|S|}\} \subseteq \{p_1,\ldots,p_K\}$ and counts $c \in \mathbb{Z}_{\geq 0}^{|S|}$ with $\sum_j c_j = N$
\State Let $\{i_1,\ldots,i_{|S|}\} \gets \{i : w_i > 0\}$;\; $s_j \gets p_{i_j}$,\; $\tilde w_j \gets w_{i_j}$;\; $S \gets \{s_1,\ldots,s_{|S|}\}$
\State $c_j \gets \lfloor N\,\tilde w_j \rfloor$ for all $j$
\For{$N - \sum_j c_j$ steps}\Comment{greedy fill}
  \State $j^\star \gets \arg\min_j \bigl\| q - \sum_\ell \tfrac{c_\ell + \mathbf{1}[\ell=j]}{N}\, s_\ell \bigr\|_2^2$;\quad $c_{j^\star} \gets c_{j^\star} + 1$
\EndFor
\For{$t = 1$ \textbf{to} $T$}\Comment{local-swap refinement}
  \State improved $\gets$ false
  \For{each $(j,k)$ with $c_j > 0$, $j \neq k$}
    \If{swapping one unit from $j$ to $k$ strictly reduces \cref{eq:integerize}}
      \State $c_j \gets c_j - 1$;\; $c_k \gets c_k + 1$;\; improved $\gets$ true
    \EndIf
  \EndFor
  \If{not improved} \textbf{break} \EndIf
\EndFor
\State \Return $S, c$
\end{algorithmic}
\end{algorithm}
\textbf{Frank--Wolfe support set.}
From the preselected pool of $K$ candidates, we solve the approximate Carath\'{e}odory problem by running the Frank--Wolfe (conditional gradient) algorithm~\cite{frank1956algorithm, jaggi2013revisiting} on the $\ell_2$-squared objective over the probability simplex (\cref{alg:fw}).
Selection reduces to finding a sparse $w\in\Delta^K$ whose convex combination $Pw$ approximates $q$:
\begin{equation}
  \min_{w\in\Delta^K}\; \|q - Pw\|_2^2.
  \label{eq:fw_obj}
\end{equation}
The support of $w$ is the selected subset: $p_i$ is picked iff $w_i > 0$.
The method is projection-free, requires only inner-product evaluations, and its iterates are sparse convex combinations by design~\cite{clarkson2010coresets_fw}.

Starting from the single vertex $w = e_{v^\star}$, where $v^\star = \arg\max_{i\in[K]} \langle q, p_i\rangle$, each iteration forms the residual $r = q - Pw$, picks the vertex index $v = \arg\max_{i\in[K]} \langle r, p_i\rangle$, and takes an exact line search along the edge from $w$ to $e_v$.
A step adds at most one new nonzero to $w$, so iterates stay sparse throughout, and near-duplicates barely move the residual and are naturally skipped.
We stop as soon as $\|q - Pw\|_2^2 \leq \varepsilon$ or $|\{i : w_i > 0\}| = m$, whichever comes first.
The cap $m$ is the binding constraint in practice: left alone, FW keeps adding vertices and $w$ eventually fills in over most of the pool.
The algorithm is defined in the appendix, \cref{sec:app_core_algorithms} as \cref{alg:fw}; we spell out Frank--Wolfe for completeness. 

\textbf{Geometric integerization.}
Frank--Wolfe (\cref{alg:fw}) returns fractional weights, which cannot be used directly for finetuning.
Training on a uniformly weighted subset can outperform loss reweighting, which is prone to variance amplification and overfitting to noisy samples~\cite{byrd2019effect}.
We therefore convert the fractional weights into integer counts that define an equally weighted multiset, preserving the geometric approximation quality (\cref{alg:integerize}).

Write the indices of the nonzero entries of $w$ as $\{i_1,\ldots,i_{|S|}\} := \{j \mid w_j > 0\}$, let $S := \{s_1,\ldots,s_{|S|}\} = \{ p_j \mid w_j >0 \}$ be the corresponding chosen points from \cref{alg:fw}, and for every $j\in \{1,\ldots, |S|\}$, let $\tilde w_j := w_{i_j}$.
We look for integer counts $c\in\mathbb{Z}_{\geq 0}^{|S|}$ with $\sum_j c_j = N$ minimizing the reconstruction error
\begin{equation}
  \Bigl\| q - \sum_{j=1}^{|S|} \tfrac{c_j}{N}\, s_j \Bigr\|_2^2,
  \label{eq:integerize}
\end{equation}
i.e.\ the $N$-point multiset over $S$ whose uniform mean is closest to $q$.
Turning the continuous weights into discrete counts is tricky because the rounded multiset must satisfy two constraints at once: it must contain exactly $N$ examples, and its empirical mean should remain close to the Frank--Wolfe convex combination.
We therefore view integerization as a small discrete reconstruction problem around the continuous solution.
The three steps below progressively restore feasibility and reduce the same geometric objective in \cref{eq:integerize}:
\begin{enumerate}
  \item \textbf{Floor allocation:} We start by setting $c_j = \lfloor N\tilde w_j \rfloor$. This gives each point the number of copies suggested by its Frank--Wolfe weight, while making sure we do not exceed the budget. Very small-weight points naturally get zero copies at this stage.
  \item \textbf{Greedy fill:} Rounding down leaves a few unallocated copies. We add them one at a time to the point that most reduces the reconstruction error in \cref{eq:integerize}, so the leftover budget is used to repair the geometric shift introduced by flooring.
  \item \textbf{Local-swap refinement:} Finally, we check whether moving one copy from one support point to another improves the reconstruction. This keeps the budget fixed, but corrects small rounding mistakes left by the greedy fill. We run this pairwise check for two passes.
\end{enumerate}
After this, each support point $s_j$ is included $c_j$ times to form the final multiset of exactly $N$ sequences.

\textbf{Full selection algorithm.}
Combining both steps, the selection pipeline (\cref{alg:select}) takes the candidate pool and budget and returns the final training multiset.

\begin{algorithm}[t]
\caption{\textsc{HullFTSelect}$(q, \{p_1,\ldots,p_K\}, N, \varepsilon, m, T)$}
\label{alg:select}
\begin{algorithmic}[1]
\Statex \textbf{Input:} query $q$, candidate pool $\{p_1,\ldots,p_K\}$, budget $N$, FW tolerance $\varepsilon$, support cap $m$, swap passes $T$
\Statex \textbf{Output:} support points $S$, counts $c$
\State $w \gets \textsc{FrankWolfe}(q, \{p_1,\ldots,p_K\}, \varepsilon, m)$
\State $S,\, c \gets \textsc{Integerize}(q, \{p_1,\ldots,p_K\}, w, N, T)$
\State \Return $S,\, c$
\end{algorithmic}
\end{algorithm}

\subsection{Efficient finetuning via gradient reuse}

The integerized multiset contains repeated sequences: a support point $s_j$ with count $c_j$ appears $c_j$ times in the finetuning batch.
A na\"{i}ve implementation treats these copies as independent samples and performs a separate forward-backward pass for each one, causing the computational cost to scale linearly with $N$ regardless of duplication.
We exploit the integer multiplicities produced by geometric integerization to amortize gradient computation across repeated optimizer steps (\cref{alg:grad_reuse}).
Rather than recomputing the gradient at every repeated step on $s_j$, we \emph{cache and reuse gradients}: we refresh $\tilde{g}$ once every $r$ steps and apply the cached value for the intermediate $r{-}1$ updates.
Let $t = 0, 1, \ldots, c_j - 1$ index the repeated steps on $s_j$:
\begin{equation}
  \tilde{g}_t =
  \begin{cases}
    \nabla_\theta \mathcal{L}(\theta_t;\, s_j) & \text{if } t \bmod r = 0, \\
    \tilde{g}_{t-1} & \text{otherwise,}
  \end{cases}
  \qquad \theta_{t+1} = \textsc{AdamStep}(\theta_t,\, \tilde{g}_t,\, \eta).
  \label{eq:grad_reuse}
\end{equation}
This reduces forward-backward passes from $N$ to $\lceil N/r \rceil$.
For small refresh intervals $r$, the parameter drift between consecutive steps on identical input is mild enough that reusing a cached gradient preserves convergence quality (see \cref{sec:expts}).
The full gradient-reuse procedure is defined in \cref{sec:app_core_algorithms} as \cref{alg:grad_reuse}. 
For gradient reuse to work, all copies of a given sequence must appear \emph{consecutively} in the training order.
Integerization gives this property by construction: multiplicities are fixed upfront, so each unique text is trained for all its $c_j$ steps in a single consecutive block.
We test whether this structure can be imposed on SIFT in \cref{sec:res_sift_reuse}.

\section{Experimental results}
\label{sec:expts}
\textbf{The setup. }We evaluate on 12 subsets of The Pile~\cite{gao2020pile} using GPT-2~\cite{radford2019gpt2}. For each subset, we average the results over 150 test queries. 
All methods share a $K{=}200$ nearest-neighbor candidate pool retrieved via a precomputed FAISS index. 
Following~\cite{hubotter2025sift}, we use the normalized RoBERTa encoder for embeddings and finetune with Adam ($lr{=}5{\times}10^{-5}$). 
Experiments run on a single NVIDIA A100 GPU, except for the CPU evaluation in \cref{sec:cpu_test}. 
Our metric is Bits-Per-Byte (BPB\%) relative to a non-finetuned baseline. 
Since TTFT happens at inference time, we report the BPB\% achievable within a given \emph{total-runtime budget} $T$ (selection + finetuning), sweeping $N \in [1, 50]$. For more details, see \cref{sec:app_experimental_protocol} and \cref{sec:app_ablations} for extended experiments and ablations, which cover FW and Carath\'{e}odory selection variants, kNN candidate pool sizing, and qualitative selection comparisons.

\begin{figure}[t]
\centering
\includegraphics[width=.9\linewidth]{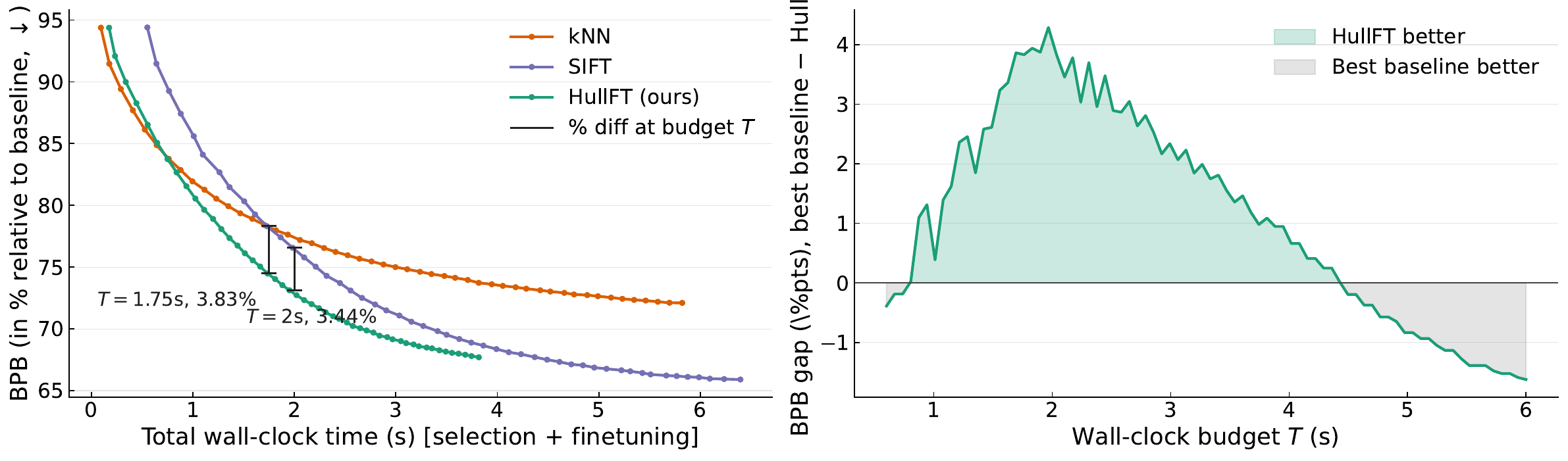}
\caption{\emph{Left}: BPB\% vs total runtime (selection + finetuning) as we sweep $N\!\in\![1,50]$ for each method.
Our method (green) is Pareto-dominant for every budget $T\!\lesssim\!4s$.
Vertical lines at $T{=}1.75s$ and $T{=}2.0s$ mark the BPB\% gap to the best baseline: our method is $3.83\%$ and $3.44\%$ lower than the best baseline at those budgets, respectively.
\emph{Right}: Quality gap (best-baseline BPB\% $-$ ours) in $\%$ as a function of the total-runtime budget $T$, where the best baseline is the lower-BPB\% method among kNN and SIFT at that budget.
Positive values (green) mean our method wins at that budget; negative (grey) means the best baseline wins.
Our method dominates for every $T\!\leq\!4.5s$\, with largest margins at tight budgets, which is exactly the regime a latency-sensitive deployment cares about.}
\label{fig:avg_improv}
\end{figure}

\textbf{Quality--runtime tradeoff.}
We compare HullFT against SIFT\cite{hubotter2025sift}, which selects examples to reduce response uncertainty by optimizing information gain while accounting for information duplication, and kNN\cite{hardt2024ttt_nn}, which finetunes on top-$N$ nearest neighbors, across a range of total-runtime budgets $T$.
\Cref{fig:avg_improv} reports the average BPB\% across all datasets as a function of total runtime, together with the BPB\% gap between HullFT and the best baseline available at each budget.
\Cref{tab:per_subset_budgets} reports the corresponding per-subset BPB\% values at the two representative budgets $T{=}1.75s$ and $T{=}2.0s$.
Across every total-runtime budget $T\!\leq\!4s$\,, our method achieves a strictly lower BPB\% than both SIFT and kNN.
The gap over the best baseline is largest at tight budgets: $\mathbf{6.4\%}$ lower at $T{=}0.75s$, $\mathbf{3.8\%}$ lower at $T{=}1.75s$, $\mathbf{3.4\%}$ lower at $T{=}2.0s$, and closes only once our method saturates its $N$ budget near $3.8s$.
In the latency-critical TTFT regime, where every additional second of total runtime is visible in user-facing latency, our method is Pareto-dominant.
\Cref{fig:avg_improv} (right) plots this best-baseline gap continuously as a function of $T$: our method dominates the strongest competing baseline for every $T\!\lesssim\!4.5s$ and the margin grows as the budget tightens, exactly the regime a latency-sensitive deployment cares about.

\begin{figure}[!htbp]
\centering
\includegraphics[width=.9\linewidth]{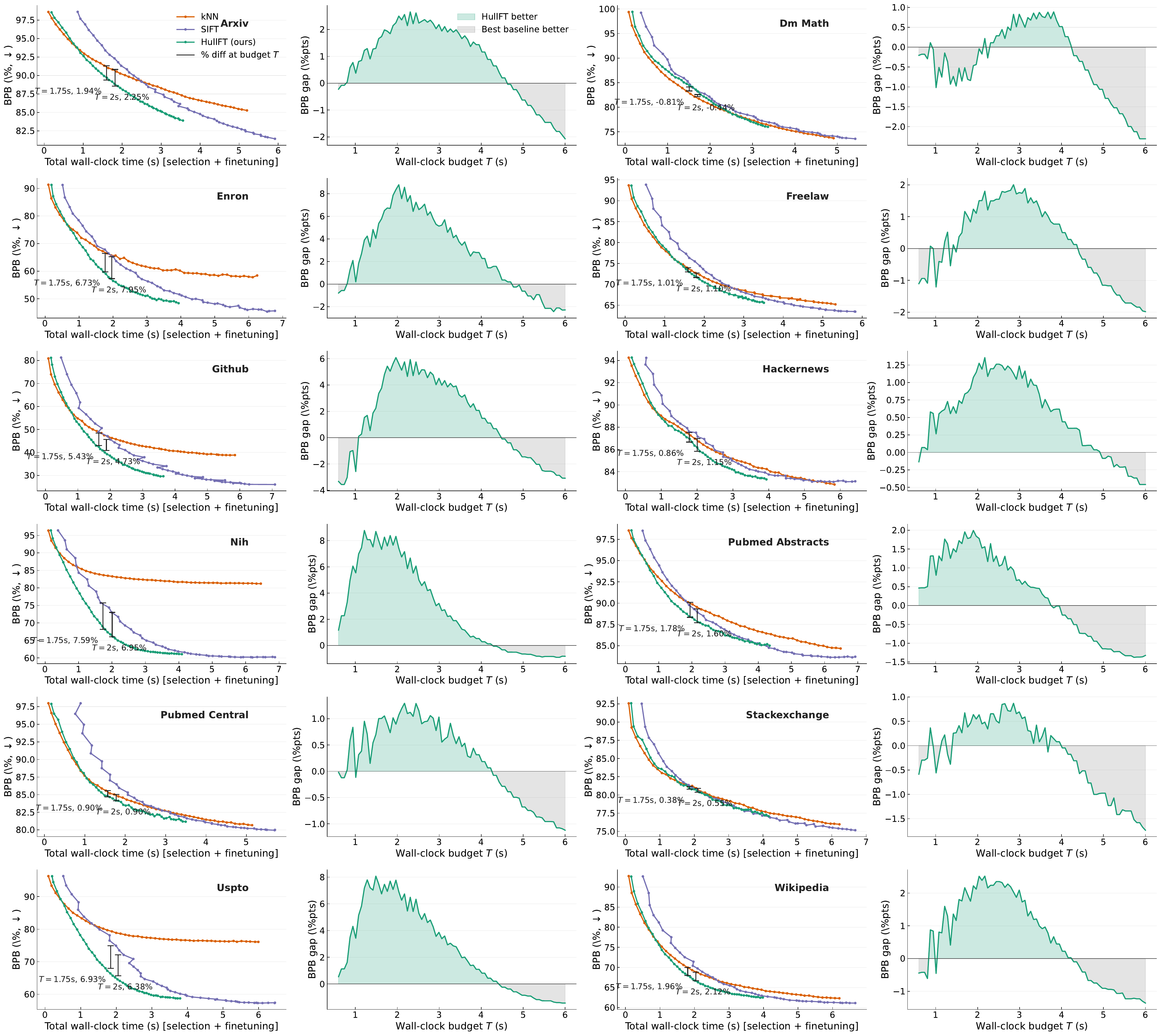}
\caption{Per-subset breakdown for all 12 Pile subsets (2 subsets per row).
For each subset we show, \emph{left}: BPB\% vs.\ wall-clock time, and \emph{right}: the quality gap (best baseline $-$ HullFT, where the best baseline is the better of kNN and SIFT) as a function of the total-runtime budget $T$. }
\label{fig:per_subset_all}
\end{figure}

\begin{table}[t]
\caption{Per-subset BPB\% at time budgets $T{=}1.75s$ (left) and $T{=}2.0s$ (right), using each method's average selection size. $\Delta$ shows the difference between the best baseline (kNN or SIFT) and HullFT.}
\label{tab:per_subset_budgets}
\centering
\footnotesize
\setlength{\tabcolsep}{3.5pt}
\noindent\begin{minipage}[t]{0.45\linewidth}
\centering
\textbf{$T{=}1.75s$} \\[0.3em]
\begin{tabular}{@{}l c c c >{\columncolor{green!10}}c@{}}
\toprule
\textbf{Subset} & \textbf{kNN} & \textbf{SIFT} & \textbf{Ours} & \textbf{$\Delta$} \\
\midrule
ArXiv            & $91.33$ & $91.80$ & $\mathbf{89.39}$ & $\mathbf{1.94}\downarrow$ \\
DM Math.\        & $\mathbf{83.31}$ & $85.29$ & $84.12$ & \cellcolor{red!10} $\mathbf{-0.81}\uparrow$ \\
Enron            & $66.47$ & $67.84$ & $\mathbf{59.74}$ & $\mathbf{6.73}\downarrow$ \\
FreeLaw          & $73.99$ & $76.32$ & $\mathbf{72.98}$ & $\mathbf{1.01}\downarrow$ \\
GitHub           & $48.42$ & $48.95$ & $\mathbf{42.98}$ & $\mathbf{5.44}\downarrow$ \\
HackerNews       & $87.52$ & $87.92$ & $\mathbf{86.66}$ & $\mathbf{0.86}\downarrow$ \\
NIH              & $83.46$ & $75.72$ & $\mathbf{68.13}$ & $\mathbf{7.59}\downarrow$ \\
PubMed Abs.\     & $90.11$ & $90.27$ & $\mathbf{88.33}$ & $\mathbf{1.78}\downarrow$ \\
PubMed Cent.\    & $85.57$ & $86.47$ & $\mathbf{84.67}$ & $\mathbf{0.90}\downarrow$ \\
StackEx.\        & $81.16$ & $81.74$ & $\mathbf{80.78}$ & $\mathbf{0.38}\downarrow$ \\
USPTO            & $79.69$ & $74.92$ & $\mathbf{67.99}$ & $\mathbf{6.93}\downarrow$ \\
Wikipedia        & $69.96$ & $72.54$ & $\mathbf{68.01}$ & $\mathbf{1.95}\downarrow$ \\
\midrule
\textbf{Avg.}   & $78.42$ & $78.31$ & $\mathbf{74.48}$ & $\mathbf{3.83}\downarrow$ \\
\bottomrule
\end{tabular}
\end{minipage}\hfill\begin{minipage}[t]{0.45\linewidth}
\centering
\textbf{$T{=}2.0s$} \\[0.3em]
\begin{tabular}{@{}l c c c >{\columncolor{green!10}}c@{}}
\toprule
\textbf{Subset} & \textbf{kNN} & \textbf{SIFT} & \textbf{Ours} & \textbf{$\Delta$} \\
\midrule
ArXiv            & $90.82$ & $90.84$ & $\mathbf{88.57}$ & $\mathbf{2.25}\downarrow$ \\
DM Math.\        & $\mathbf{82.14}$ & $83.97$ & $82.58$ & \cellcolor{red!10} $\mathbf{-0.44}\uparrow$ \\
Enron            & $65.67$ & $65.26$ & $\mathbf{57.32}$ & $\mathbf{7.94}\downarrow$ \\
FreeLaw          & $72.73$ & $74.40$ & $\mathbf{71.63}$ & $\mathbf{1.10}\downarrow$ \\
GitHub           & $46.97$ & $45.61$ & $\mathbf{40.88}$ & $\mathbf{4.73}\downarrow$ \\
HackerNews       & $86.98$ & $87.51$ & $\mathbf{85.83}$ & $\mathbf{1.15}\downarrow$ \\
NIH              & $83.08$ & $73.01$ & $\mathbf{66.05}$ & $\mathbf{6.95}\downarrow$ \\
PubMed Abs.\     & $89.54$ & $89.30$ & $\mathbf{87.70}$ & $\mathbf{1.60}\downarrow$ \\
PubMed Cent.\    & $85.03$ & $85.31$ & $\mathbf{84.13}$ & $\mathbf{0.90}\downarrow$ \\
StackEx.\        & $80.91$ & $81.03$ & $\mathbf{80.37}$ & $\mathbf{0.54}\downarrow$ \\
USPTO            & $78.86$ & $72.08$ & $\mathbf{65.70}$ & $\mathbf{6.38}\downarrow$ \\
Wikipedia        & $68.83$ & $70.41$ & $\mathbf{66.71}$ & $\mathbf{2.12}\downarrow$ \\
\midrule
\textbf{Avg.}   & $77.63$ & $76.56$ & $\mathbf{73.12}$ & $\mathbf{3.44}\downarrow$ \\
\bottomrule
\end{tabular}
\end{minipage}
\end{table}
\textbf{Per-subset breakdown.}
We test whether the aggregate runtime advantage is driven by a few favorable subsets or appears broadly across The Pile.
\Cref{fig:per_subset_all} shows the full per-subset quality--runtime curves and best-baseline gaps.
\Cref{tab:per_subset_budgets} shows per-subset BPB\% for kNN, SIFT, and HullFT at $T{=}1.75s$ (kNN $N^{\star}{=}15$, SIFT $N^{\star}{=}11$, HullFT $N^{\star}{=}19$) and at $T{=}2.0s$ (kNN $N^{\star}{=}17$, SIFT $N^{\star}{=}13$, HullFT $N^{\star}{=}22$), using the $N^{\star}$ each method attains in our sweep for that time budget.
At $T{=}1.75s$, our method improves over the best performing baseline on 11 of 12 subsets, with an average margin of $3.83\%$; at $T{=}2.0s$ the same holds, with an average margin of $3.44\%$ (as in \cref{fig:avg_improv}).
The quality gap is largest on subsets where finetuning signal continues to improve with larger multisets (Enron, GitHub, USPTO, NIH), which is exactly where our method's ability to reach a higher $N^{\star}$ within the same time budget matters most.
The mechanism is a two-stage speedup: our selection stage is $\mathbf{8.8\times}$ faster than SIFT at $N{=}50$ ($0.059s$ vs $0.524s$, the saturation point our method reaches near $T{\approx}3.8s$) and over $\mathbf{12\times}$ faster averaged across $N\!\in\![1,50]$, and Gradient Reuse further cuts finetuning by $\mathbf{1.48\times}$ (\cref{tab:gr_ablation}).
These speedups compound so that, inside a given time budget, our method can afford a substantially larger $N^{\star}$, and the extra multiset mass more than offsets matched-$N$ quality differences relative to the baselines.
Against kNN at matched $N$ (\cref{fig:n_select_bpb}), our BPB\% is lower at every $N\!\geq\!10$: pure nearest-neighbour retrieval saturates because its top-$k$ is redundant under the query embedding, whereas the convex-hull support set keeps adding signal.

\begin{wrapfigure}[16]{r}{0.45\linewidth}
\vspace{-1em}
\centering
\includegraphics[width=\linewidth]{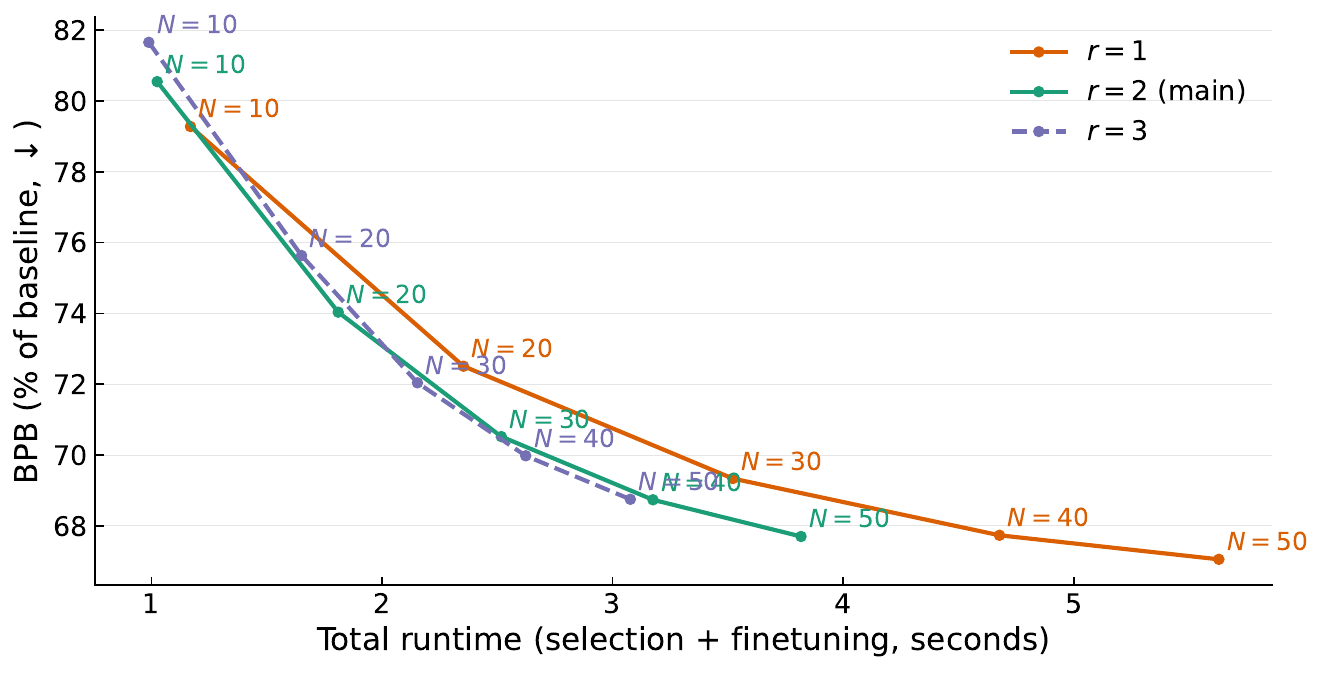}
\caption{BPB\% vs.\ total runtime (selection + finetuning) for HullFT at gradient-reuse depths $r\in\{1,2,3\}$.
Points correspond to $N\in\{10,20,30,40,50\}$.
At any given runtime budget, $r{=}2$ matches or beats $r\in\{1,3\}$, saves wall time at a modest BPB\% cost.}
\label{fig:abl_r}
\vspace{-1em}
\end{wrapfigure}
\textbf{Gradient reuse ablation.}
\Cref{tab:gr_ablation} isolates the effect of Gradient Reuse on our method at $N{=}50$.
Averaged over 12 subsets, GR yields a $\mathbf{1.48\times}$ finetuning speedup at a cost of only $0.64\%$ of average BPB\%.
On subsets where integerization produces few repeats (GitHub, Enron) GR is effectively a no-op both in runtime and quality, while on subsets with many repeats the speedup is substantially larger.
This is what enables the compounded savings that \cref{fig:avg_improv} and \cref{tab:per_subset_budgets} rely on.
Appendix \cref{sec:app_ablations} shows the BPB-versus-$N$ comparison across all three methods for $N \in \{1,\dots,50\}$. 
\Cref{fig:abl_r} ablates the gradient refresh interval $r$, where $r{=}1$ is the standard finetuning without GR; $r{=}2$ already delivers a substantial runtime reduction at a very low BPB\% cost, while $r{=}3$ saves further wall time at a more noticeable quality cost. We adopt $r{=}2$ as our default for the best speedup--accuracy tradeoff.

\begin{table}[t]
\centering
\caption{Ablation: our method with and without Gradient Reuse (at $r{=}2$) at $N{=}50$.
GR yields a $\mathbf{1.48\times}$ average finetuning speedup at a cost of $0.64\%$ average BPB\%; on subsets where the
integerized multiset has few repeats (GitHub, Enron) GR is a no-op both in time and quality, while
on subsets with more repeats the speedup is larger.}
\label{tab:gr_ablation}
\vspace{0.4em}
\setlength{\tabcolsep}{2.5pt}
\resizebox{\textwidth}{!}{%
\begin{tabular}{ll cccccccccccc >{\columncolor{green!10}}c}
\toprule
& & \textbf{ArXiv} & \textbf{DM Math} & \textbf{Enron} & \textbf{FreeLaw} & \textbf{GitHub} & \textbf{HackerNews} & \textbf{NIH} & \textbf{PubMed Abs.} & \textbf{PubMed Central} & \textbf{StackExchange} & \textbf{USPTO} & \textbf{Wikipedia} & \textbf{Average} \\
\midrule
\multirow{2}{*}{\textbf{BPB\% $\downarrow$}}
 & w/o & $83.19$ & $74.08$ & $48.70$ & $64.32$ & $29.64$ & $82.83$ & $60.71$ & $84.42$ & $80.59$ & $76.28$ & $58.08$ & $61.82$ & $67.06$ \\
 & w/  & $83.90$ & $76.00$ & $48.42$ & $65.58$ & $29.64$ & $83.31$ & $61.08$ & $85.00$ & $81.16$ & $77.08$ & $58.72$ & $62.52$ & $67.70$ \\
\cmidrule(lr){1-15}
\multirow{2}{*}{\textbf{FT (s)}}
 & w/o & $5.40$ & $5.04$ & $3.87$ & $5.47$ & $3.58$ & $6.03$ & $6.62$ & $6.62$ & $5.31$ & $6.38$ & $6.16$ & $6.37$ & $5.57$ \\
 & w/  & $3.49$ & $3.30$ & $3.88$ & $3.45$ & $3.60$ & $3.87$ & $4.04$ & $4.28$ & $3.43$ & $4.10$ & $3.75$ & $3.92$ & $3.76$ \\
\cmidrule(lr){1-15}
\multicolumn{2}{l}{\textbf{Speedup}}
       & $\mathbf{1.54\times}$ & $\mathbf{1.53\times}$ & $\mathbf{1.00\times}$ & $\mathbf{1.59\times}$ & $\mathbf{1.00\times}$ & $\mathbf{1.56\times}$ & $\mathbf{1.64\times}$ & $\mathbf{1.55\times}$ & $\mathbf{1.55\times}$ & $\mathbf{1.56\times}$ & $\mathbf{1.64\times}$ & $\mathbf{1.63\times}$ & $\mathbf{1.48\times}$ \\
\bottomrule
\end{tabular}%
}
\end{table}

\begin{figure}[t]
\centering

\begin{minipage}[c]{0.56\linewidth}
\centering
\includegraphics[width=\textwidth]{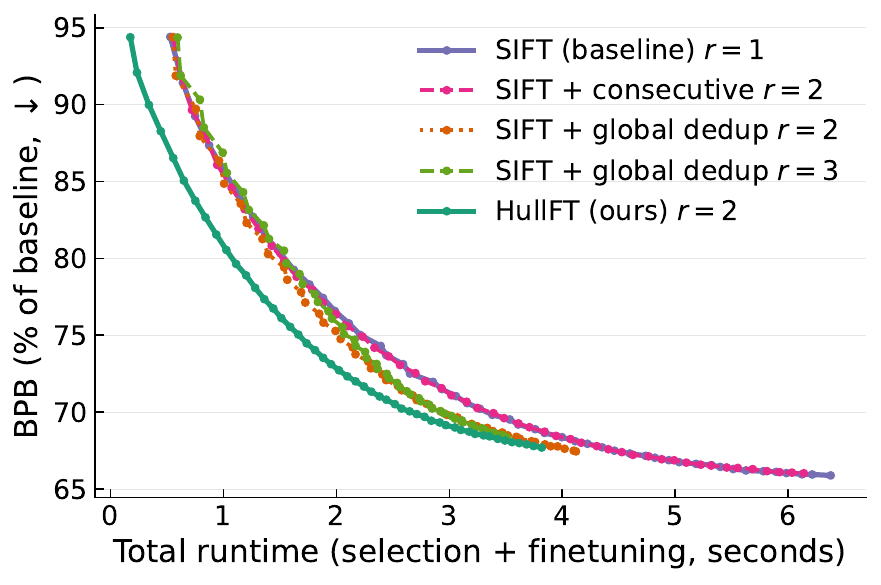}
\caption{Comparison of SIFT gradient-reuse variants across selection budgets and HullFT.}
\label{fig:sift_gr_variants}
\end{minipage}
\hfill
\begin{minipage}[c]{0.40\linewidth}
\centering
\captionof{table}{Gradient-reuse variants applied to SIFT-derived schedules at $N{=}50$, averaged over 12 Pile subsets.
Both SIFT-based reuse strategies degrade quality relative to plain SIFT, while HullFT avoids the reorder penalty because multiplicities are fixed before finetuning.}
\label{tab:sift_gr_variants}
\vspace{2mm}
{\footnotesize
\setlength{\tabcolsep}{3pt}
\begin{tabular}{@{}lccr@{}}
\toprule
\textbf{Method} & \textbf{$r$} & \textbf{BPB\%} & \textbf{Time(s)} \\
\midrule
SIFT (baseline) & 1 & 65.89 & 6.38 \\
SIFT + consecutive & 2 & 66.03 & 6.14 \\
SIFT + global dedup & 2 & 67.45 & 4.16 \\
SIFT + global dedup & 3 & 68.37 & 3.51 \\
HullFT (ours) & 2 & 67.70 & 3.82 \\
\bottomrule
\end{tabular}
}
\end{minipage}

\end{figure}

\textbf{SIFT gradient reuse experiment.}
\label{sec:res_sift_reuse}
We investigate whether Gradient Reuse can be applied to other selection methods to improve their efficiency.
While kNN retrieval~\cite{hardt2024ttt_nn} returns $N$ distinct sequences without repetitions, SIFT~\cite{hubotter2025sift} can revisit the same sequence at non-consecutive steps.
To enable Gradient Reuse for SIFT, we evaluate two modified schedules: \emph{global deduplication}, which collapses all occurrences of each unique text into a single block with aggregated counts, and \emph{consecutive grouping}, which merges only adjacent duplicates.
\Cref{fig:sift_gr_variants} reports the BPB\%--runtime behavior of these SIFT reuse variants across selection budgets, with HullFT included for reference.
\Cref{tab:sift_gr_variants} reports the same comparison at $N{=}50$, averaged over the 12 Pile subsets.

The results show that Gradient Reuse accelerates SIFT, presenting a clear quality--efficiency tradeoff.
Global deduplication provides a substantial speedup but alters the original training order, increasing the BPB\%. Consecutive grouping preserves the schedule better, yielding a smaller runtime gain with minimal quality degradation.
Unlike SIFT, HullFT inherently produces repeated contiguous blocks during integerization, allowing it to fully exploit Gradient Reuse without post-hoc reordering.

\begin{figure}[t]
\centering
\includegraphics[width=.9\linewidth]{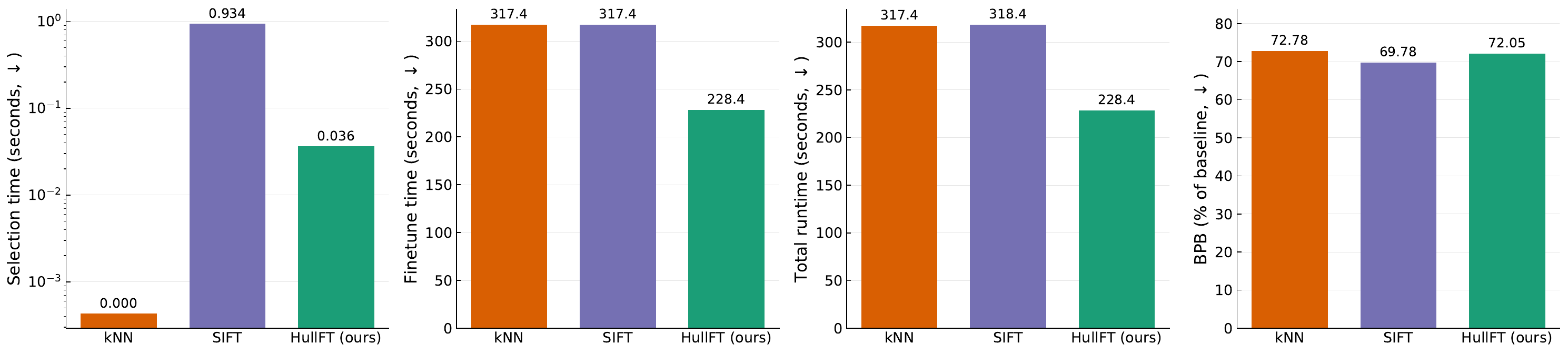}
\caption{CPU-only comparison at $N{=}20$ over 6 Pile subsets.
\textbf{Left}: mean selection time on CPU.
\textbf{Middle-left}: mean finetuning time.
\textbf{Middle-right}: total runtime (selection + finetuning).
\textbf{Right}: BPB\% relative to the non-finetuned baseline.
HullFT is $25.8\times$ faster than SIFT at selection and saves about $89s$ end-to-end, at a modest BPB\% cost relative to SIFT.}
\label{fig:cpu_test}
\end{figure}

  \textbf{CPU-only evaluation.}
\label{sec:cpu_test}
We repeat the $N{=}20$ comparison on CPU only over 6 Pile subsets (ArXiv, DM Mathematics, Enron, FreeLaw, GitHub, HackerNews).
\Cref{fig:cpu_test} reports mean CPU selection time, finetuning time, total runtime, and BPB\% for kNN, SIFT, and HullFT.
HullFT's FW selector is $\mathbf{25.8\times}$ faster than SIFT's on CPU ($0.036s$ vs.\ $0.934s$).
Because Gradient Reuse reduces the number of forward-backward passes, finetuning is also faster: $228.4s$ vs.\ $317.4s$ for kNN and $318.4s$ for SIFT, saving roughly $\mathbf{89s}$ end-to-end.
On BPB\%, HullFT ($72.05\%$) is close to kNN ($72.78\%$) and within $2.3\%$ of SIFT ($69.78\%$), while being substantially faster end-to-end (\cref{fig:cpu_test}).

\begin{figure}[t]
\centering
\includegraphics[width=0.85\linewidth]{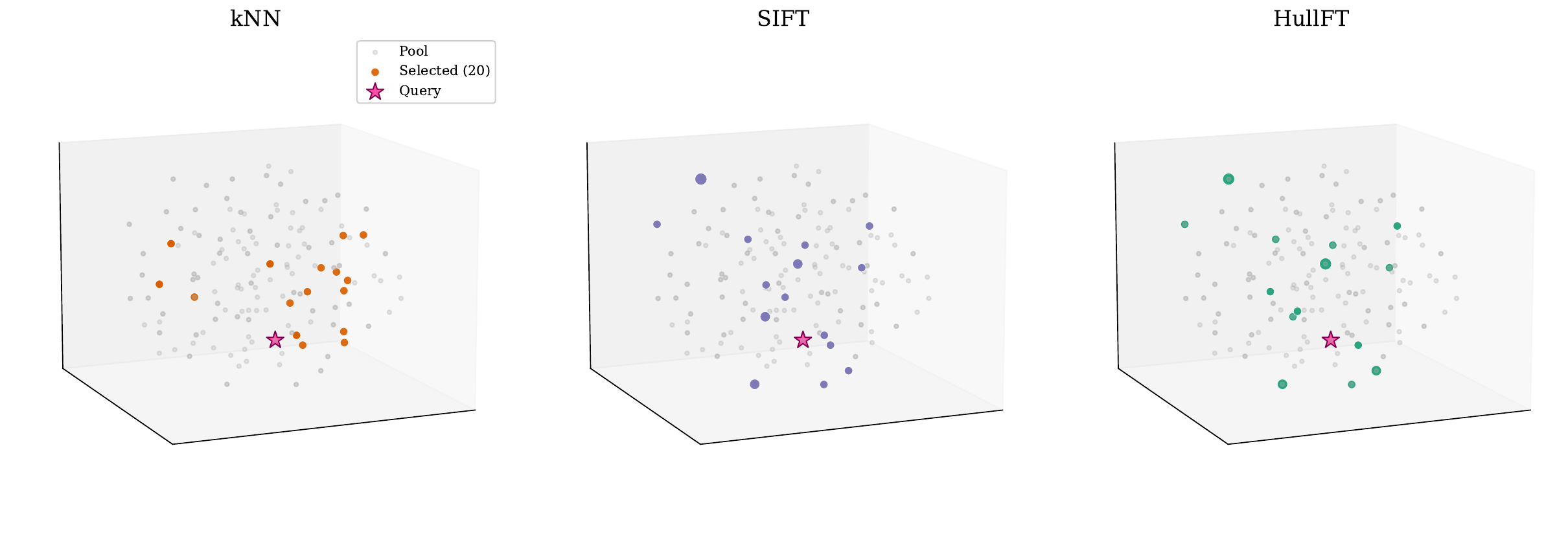}
\caption{3D t-SNE projection of the candidate pool (gray) and the $N{=}20$ points selected by kNN, SIFT, and HullFT (colored) for a representative ArXiv query (pink star). HullFT produces a similarly diverse selection to SIFT through convex geometry, while incurring substantially lower selection latency (\cref{sec:cpu_test}).}
\label{fig:arxiv_diversity}
\end{figure}

\textbf{Selection diversity.}
\Cref{fig:arxiv_diversity} visualizes, for a representative ArXiv query at $N{=}20$, the support sets chosen by kNN, SIFT, and HullFT in a 3D t-SNE projection of the candidate pool.
kNN concentrates its budget in a tight neighborhood of the query, while SIFT places its points more broadly around the query to reduce redundancy.
HullFT yields a comparably diverse support, obtained directly from the geometry of the convex approximation in \cref{sec:method} rather than from any explicit diversity term.
This diversity is achieved at a fraction of SIFT's per-query selection cost (\cref{sec:cpu_test}), consistent with our central thesis that geometric reconstruction recovers a comparably diverse selection at substantially lower latency.

\textbf{More results.} The appendix provides extended experiments and results. 
Specifically, \cref{sec:app_ablations} includes a comprehensive sweep of the budget $N$, integerization ablations, Carath\'{e}odory variants comparisons, and sensitivity analyses for $\varepsilon$, $r$, and $k$. 
We also provide per-subset breakdowns (\cref{sec:app_per_subset_vs_each}), qualitative results, and text examples (\cref{sec:app_qualitative}).

\section{Conclusion and Limitations}
\label{sec:conclusion}

We introduced \textbf{HullFT}, a test-time finetuning pipeline that frames data selection as sparse convex approximation of the query embedding over a candidate pool.
Frank--Wolfe yields a diverse, relevance-aware support set without explicit redundancy penalties, geometric integerization converts the resulting fractional weights into an exact $N$-point training multiset, and Gradient Reuse amortizes the repeated optimizer steps this multiset produces.
Across 12 subsets of The Pile, HullFT attains a strictly lower BPB\% than both kNN retrieval and SIFT at every total-runtime budget $T\!\leq\!4s$\,, with the margin widening as the budget tightens.
The gains come from a compounded two-stage speedup (selection is $12\times$ faster than SIFT on average and Gradient Reuse cuts finetuning by $1.48\times$), which lets HullFT reach a substantially larger effective $N$ inside a fixed latency budget.
These results suggest that convex geometry is a natural vehicle for latency-constrained test-time adaptation, and that the integer structure it induces opens further room for optimizer-level amortization.

\textbf{Limitations.} Similarly to SIFT~\cite{hubotter2025sift}, HullFT operates over a kNN-preselected candidate pool rather than the full corpus: the convex approximation is only as expressive as the directions present in this pool, so the quality of the support set ultimately inherits the recall ceiling of the upstream retriever and embedding model. Second, selection runs in a fixed embedding space decoupled from the base LLM: this keeps the method general and model-agnostic, but the geometry is not tailored to the loss landscape of the model being adapted, leaving potential per-model gains on the table.

\newpage
\bibliographystyle{unsrtnat}
\bibliography{hullft}

\newpage
\appendix



\section{Experimental protocol details}
\label{sec:app_experimental_protocol}

The main experiments use 12 Pile subsets and 150 test queries per subset.
This protocol is a compute-conscious version of the evaluation style used by TTT-NN~\cite{hardt2024ttt_nn} and SIFT~\cite{hubotter2025sift}, which evaluate approximately $1\%$ of the full Pile test set ($1812$ points) and report only on subsets with more than 10 test examples.
We use the larger fixed per-subset sample because our goal is not only to report a single aggregate number, but also to understand whether HullFT's runtime--quality behavior is stable across domains with different redundancy patterns.
A balanced 150-query sample gives each subset enough weight for per-domain comparisons while keeping the full sweep over methods, budgets, and ablations feasible on our available A100 compute.

We build and use precomputed per-subset retrieval pools rather than a single live global index.
This keeps the comparison focused on the TTFT components studied in the paper: query-conditioned selection from a shared candidate pool and the subsequent finetuning schedule.
All methods receive the same $K$ precomputed nearest neighbors for each query, so retrieval differences cannot explain the reported gaps.
We also exclude pool retrieval time from the main runtime measurements because it is identical across methods once the candidate pool is fixed; selection and finetuning are the parts changed by HullFT.
For context, Hardt and Sun~\cite{hardt2024ttt_nn} report mean distributed Pile retrieval time of $1.35s$ (std.\ $0.12s$).

Unless otherwise stated, all main experiments use normalized RoBERTa embeddings from \textit{socialfoundations/roberta-large-pile-lr2e-5-bs16-8gpu-1700000}, GPT-2 as the adapted model, Adam with learning rate $5{\times}10^{-5}$, $K{=}200$ preselected candidates, and selection budgets $N\in\{1,\ldots,50\}$.

\section{Algorithmic details}
\label{sec:app_algorithms}

This section provides pseudocode for the core procedures, algorithmic variants, and baselines referenced in \cref{sec:method}.
The variants modify individual stages of the HullFT pipeline.

\subsection{Core HullFT procedures}
\label{sec:app_core_algorithms}

\Cref{alg:fw} and \cref{alg:grad_reuse} give the procedures referenced in \cref{sec:method}.
Frank--Wolfe is a classical conditional-gradient method~\cite{frank1956algorithm,jaggi2013revisiting}; we include its specialization to our objective for completeness, not as a new algorithmic contribution.

\begin{algorithm}[H]
\caption{\textsc{FrankWolfe}$(q, \{p_1,\ldots,p_K\}, \varepsilon, m)$}
\label{alg:fw}
\begin{algorithmic}[1]
\State \textbf{Input:} query $q \in \R^d$, candidate pool $\{p_1,\ldots,p_K\} \subset \R^d$, tolerance $\varepsilon$, support cap $m$
\State \textbf{Output:} sparse weights $w \in \Delta^K$
\State Form $P \in \R^{d\times K}$ with $p_i$ as the $i$-th column
\State $v^\star \gets \arg\max_{i\in[K]} \langle q, p_i\rangle$;\quad $w \gets e_{v^\star}$
\While{$\|q - Pw\|_2^2 > \varepsilon$ \textbf{and} $|\{i : w_i > 0\}| < m$}
  \State $r \gets q - Pw$
  \State $v \gets \arg\max_{i\in[K]} \langle r, p_i\rangle$
  \State $\gamma^\star \gets \arg\min_{\gamma\in[0,1]} \|r - \gamma(p_v - Pw)\|_2^2$ \Comment{line search}
  \State $w \gets (1-\gamma^\star)w + \gamma^\star e_v$
\EndWhile
\State \Return $w$
\end{algorithmic}
\end{algorithm}

\begin{algorithm}[H]
\caption{\textsc{GradReuse}$(\theta_0, S, c, r, \eta)$}
\label{alg:grad_reuse}
\begin{algorithmic}[1]
\State \textbf{Input:} Base model $\theta_0$, support points $S = \{s_1,\ldots,s_{|S|}\}$, counts $c \in \mathbb{Z}_{\geq 0}^{|S|}$, refresh interval $r$, learning rate $\eta$
\State \textbf{Output:} Adapted model $\theta$
\State $\theta \gets \theta_0$
\For{$j = 1$ \textbf{to} $|S|$ with $c_j > 0$}
  \For{$t = 0$ \textbf{to} $c_j - 1$}
    \If{$t \bmod r = 0$}
      \State $\tilde{g} \gets \nabla_\theta \mathcal{L}(\theta;\, s_j)$ \Comment{forward-backward pass}
    \EndIf
    \State $\theta \gets \textsc{AdamStep}(\theta,\, \tilde{g},\, \eta)$
  \EndFor
\EndFor
\State \Return $\theta$
\end{algorithmic}
\end{algorithm}

\subsection{Exact Carath\'{e}odory reduction}
\label{sec:app_caratheodory}

The exact Carath\'{e}odory theorem guarantees that any point in the convex hull of a set in $\R^d$ can be expressed as a convex combination of at most $d+1$ points.
\Cref{alg:caratheodory} implements the standard constructive procedure: it iteratively finds an affine dependency among the current support points and eliminates the one that drives a weight to zero, repeating until the size bound $d+1$ is met.

\begin{algorithm}[H]
\caption{\textsc{Carath\'{e}odory}$(\{p_1,\ldots,p_k\},\, w)$}
\label{alg:caratheodory}
\begin{algorithmic}[1]
\State \textbf{Input:} Set of points $\{p_1, \ldots, p_k\} \subset \R^d$, weights $w \in \Delta^k$
\State \textbf{Output:} Support points $S \subseteq \{p_1, \ldots, p_k\}$, new weights $w'$ with $|S| \le d+1$
\State $A \gets \{1, \ldots, k\}$ \Comment{active index set}
\While{$|A| > d + 1$}
  \State Find non-trivial coefficients $(\alpha_i)_{i \in A}$ such that $\sum_{i \in A} \alpha_i\, p_i = 0$ and $\sum_{i \in A} \alpha_i = 0$
  \State $i^\star \gets \arg\min_{i \in A\, :\, \alpha_i > 0} \frac{w_i}{\alpha_i}$;\quad $\gamma \gets w_{i^\star} / \alpha_{i^\star}$
  \State $w_i \gets w_i - \gamma \alpha_i$ for all $i \in A$
  \State $A \gets A \setminus \{i^\star\}$
\EndWhile
\State $S \gets \{p_i : i \in A\}$;\; $w' \gets (w_i)_{i \in A}$
\State \Return $S,\, w'$
\end{algorithmic}
\end{algorithm}

\Cref{fig:abl_caratheodory} (see \cref{sec:app_abl_caratheodory}) compares exact Carath\'{e}odory selection against HullFT with different integerization strategies.
The key finding is that geometric integerization is essential: padding Carath\'{e}odory weights directly is unstable as $N$ grows, while FW with geometric integerization (HullFT) matches or beats the best Carath\'{e}odory variant at lower selection cost.

\subsection{PCA-reduced convex approximation}
\label{sec:app_pca}

When the embedding dimension $d$ is large, inner-product evaluations inside Frank--Wolfe dominate selection cost.
\Cref{alg:pca_fw} mitigates this by projecting the query and candidate pool into a lower-dimensional subspace via PCA before running the geometric selection, then mapping the chosen support back to the original space for integerization.

\begin{algorithm}[H]
\caption{\textsc{PCA\_Selection}$(q, \{p_1,\ldots,p_K\}, N, \varepsilon, m, T, d')$}
\label{alg:pca_fw}
\begin{algorithmic}[1]
\State \textbf{Input:} Query $q$, candidate pool $\{p_1,\ldots,p_K\}$, budget $N$, FW tolerance $\varepsilon$, support cap $m$, swap passes $T$, target dimension $d'$
\State \textbf{Output:} Support points $S$, integer counts $c$
\State Fit PCA on $\{p_1,\ldots,p_K\} \cup \{q\}$ to reduce dimensionality to $d'$
\State $q' \gets \textsc{Transform}(q)$;\quad $p_i' \gets \textsc{Transform}(p_i)$ for $i = 1,\ldots,K$
\State $w \gets \textsc{FrankWolfe}(q', \{p_1',\ldots,p_K'\}, \varepsilon, m)$ \Comment{or \textsc{Carath\'{e}odory} via \cref{alg:caratheodory}}
\State $S',\, c \gets \textsc{Integerize}(q', \{p_1',\ldots,p_K'\}, w, N, T)$
\State Map $S'$ back to the original pool: $S \gets \{p_i : p_i' \in S'\}$ (preserving order)
\State \Return $S,\, c$
\end{algorithmic}
\end{algorithm}

\subsection{Weight-proportional padding}
\label{sec:app_pad}

As a simpler baseline to the full geometric integerization (\cref{alg:integerize}), \cref{alg:pad_weights} allocates the discrete budget $N$ proportionally to the fractional weights via largest-remainder rounding, without the subsequent local-swap refinement step.

\begin{algorithm}[H]
\caption{\textsc{PadByWeights}$(w, N)$}
\label{alg:pad_weights}
\begin{algorithmic}[1]
\State \textbf{Input:} Fractional weights $w \in \Delta^k$, budget $N$
\State \textbf{Output:} Integer counts $c \in \mathbb{Z}_{\geq 0}^k$ such that $\sum_i c_i = N$
\State $c_i \gets \lfloor N w_i \rfloor$ for all $i \in \{1, \ldots, k\}$
\State $\rho \gets N - \sum_i c_i$ \Comment{remaining budget to allocate}
\State Let $I$ be the list of indices sorted descending by fractional remainder $N w_i - c_i$
\For{$j = 1$ \textbf{to} $\rho$}
  \State $i \gets I[j]$
  \State $c_i \gets c_i + 1$
\EndFor
\State \Return $c$
\end{algorithmic}
\end{algorithm}

\Cref{fig:abl_fw_family} (see \cref{sec:app_abl_fw_family}) compares this simpler baseline against the full geometric integerization used by HullFT.

\section{Ablations}
\label{sec:app_ablations}

This section isolates the effect of individual design choices.
Each ablation sweeps one axis while holding the rest of the pipeline fixed to the configuration used in \cref{sec:expts}.

\begin{figure}[H]
\centering
\includegraphics[width=\linewidth]{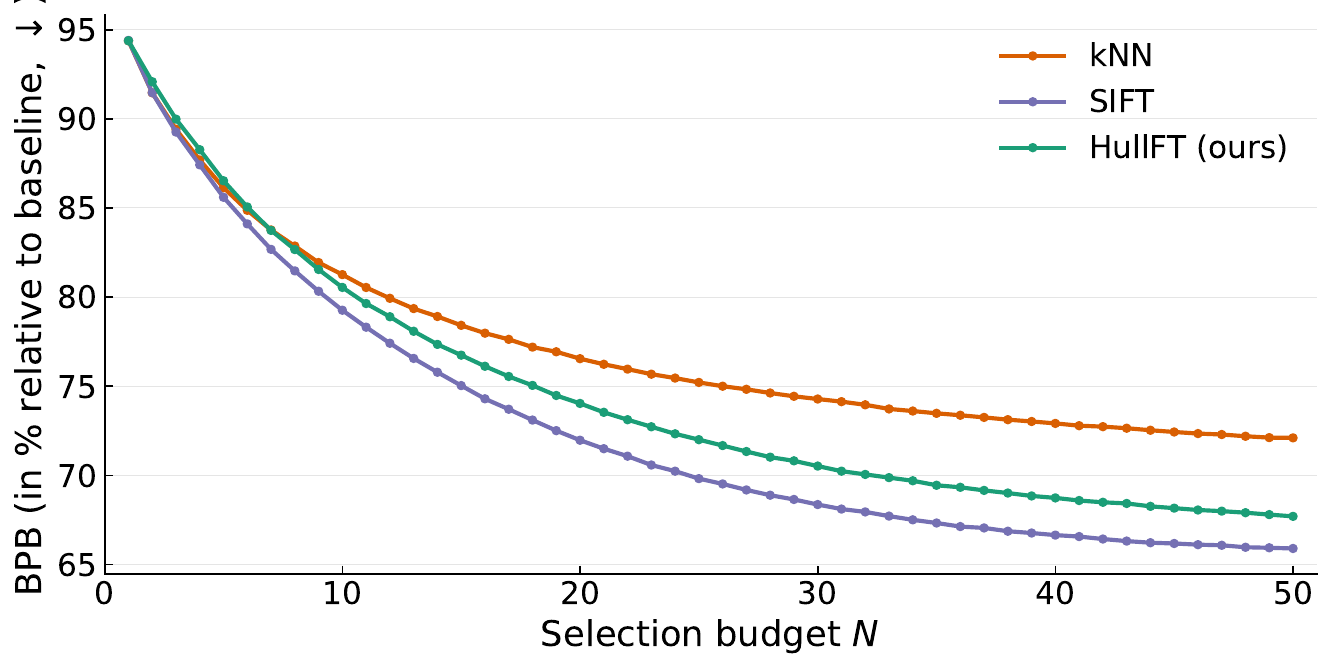}
\caption{BPB\% averaged over 12 Pile subsets as a function of selection budget $N$ for kNN, SIFT, and ours.}
\label{fig:n_select_bpb}
\end{figure}

For completeness, \cref{fig:n_select_bpb} reports BPB\% as a function of the selection budget $N$ across the three main methods.

\subsection{Integerization: pad-by-weights vs.\ geometric (FW family)}
\label{sec:app_abl_fw_family}
\begin{figure}[H]
\centering
\includegraphics[width=\linewidth]{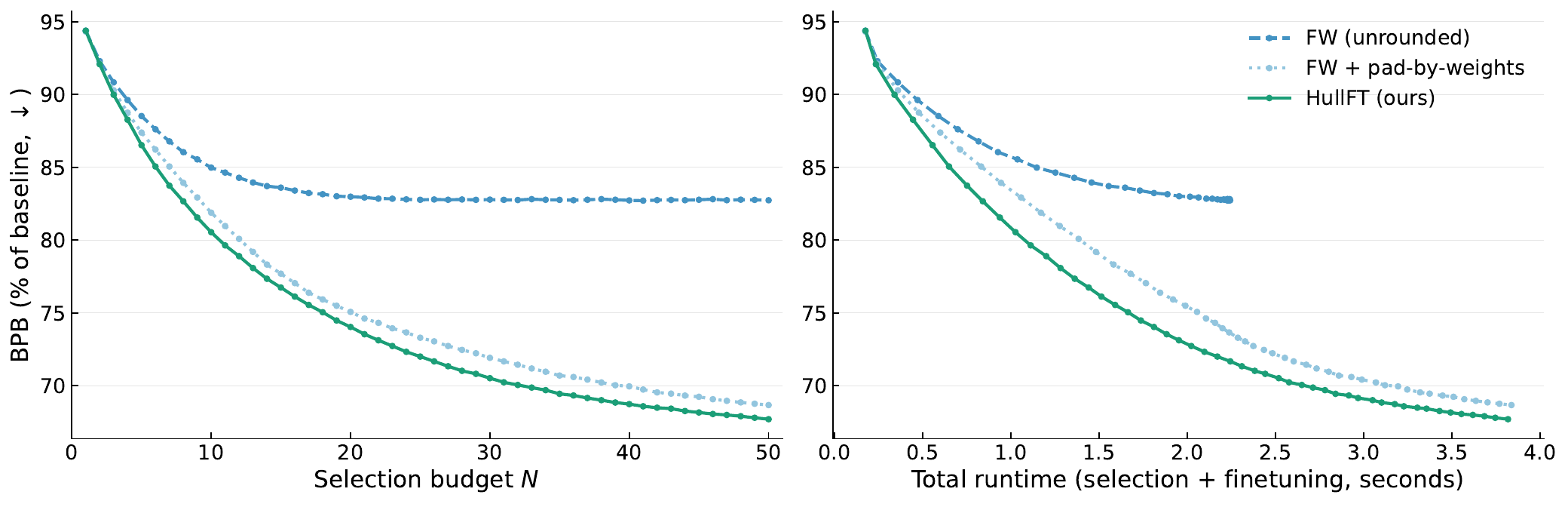}
\caption{BPB\% for the three FW integerization strategies.
\textbf{Left}: BPB\% vs.\ selection budget $N$.
\textbf{Right}: BPB\% vs.\ total runtime in seconds.}
\label{fig:abl_fw_family}
\end{figure}

We compare the fractional FW solution (no integerization), FW followed by padding by weights (\cref{alg:pad_weights}), and FW followed by the geometric integerization used by HullFT (\cref{alg:integerize}).
\Cref{fig:abl_fw_family} reports BPB\% for these three FW-based variants both as a function of selection budget $N$ and as a function of total runtime.
The right panel shows the same comparison on a total-runtime x-axis: since all three variants share the same FW selection stage, the runtime difference is negligible and quality vs time differences are the dominant effect.

\subsection{Carath\'{e}odory selection and its integerizations}
\label{sec:app_abl_caratheodory}
\begin{figure}[H]
\centering
\includegraphics[width=\linewidth]{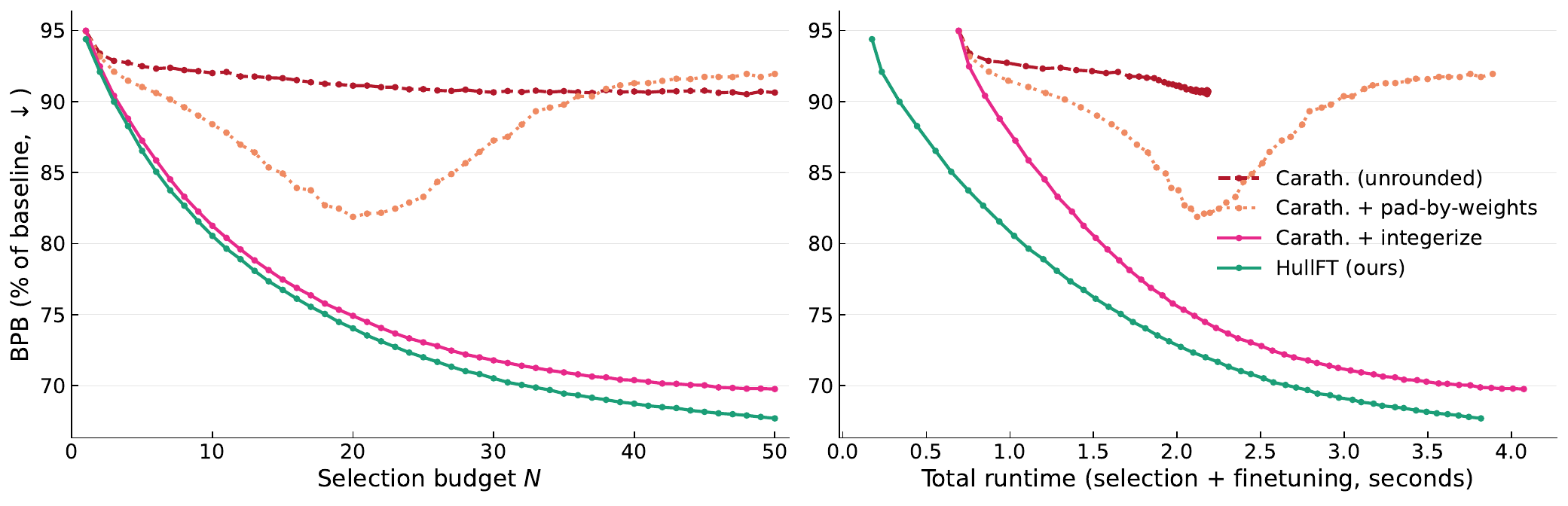}
\caption{BPB\% for Carath\'{e}odory selection (\cref{alg:caratheodory}) with three integerization strategies, with HullFT (FW + geometric integerization) plotted for reference.
\textbf{Left}: BPB\% vs.\ selection budget $N$.
\textbf{Right}: BPB\% vs.\ total runtime (selection + finetuning, seconds); the Carath\'{e}odory curves shift right relative to HullFT, reflecting its higher selection cost.}
\label{fig:abl_caratheodory}
\end{figure}

Replacing FW with exact Carath\'{e}odory (\cref{alg:caratheodory}) yields a fractional solution with at most $d{+}1$ points; we compare its pad-by-weights and geometric integerizations against HullFT.
\Cref{fig:abl_caratheodory} reports BPB\% for these Carath\'{e}odory-based variants and HullFT, plotted against both selection budget and total runtime.
Geometric integerization is essential: padding Carath\'{e}odory weights directly is unstable as $N$ grows.
The right panel makes the runtime cost of exact Carath\'{e}odory visible: its selection time is roughly $14\times$ higher than FW, so even when its BPB\% at matched $N$ is competitive, HullFT dominates on the total-runtime axis.

\subsection{Frank--Wolfe tolerance \texorpdfstring{$\varepsilon$}{epsilon}}
\label{sec:app_abl_epsilon}
\begin{figure}[H]
\centering
\includegraphics[width=\linewidth]{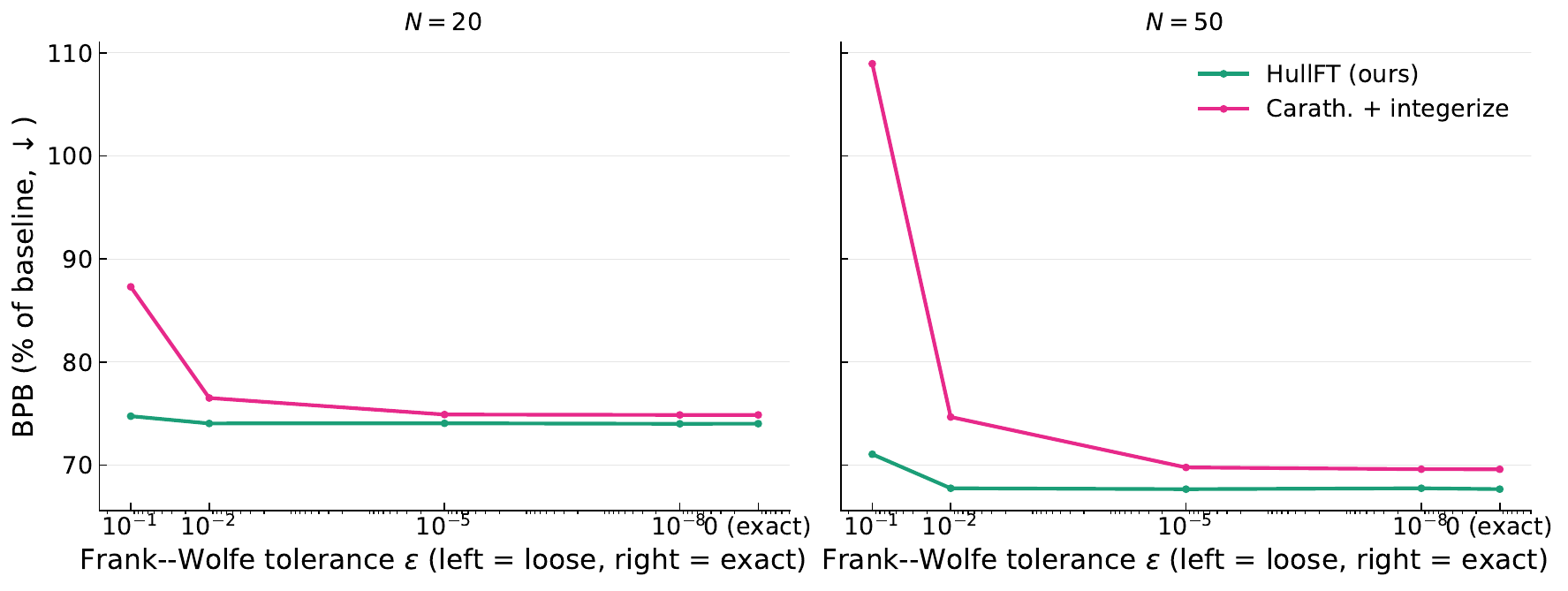}
\caption{BPB\% vs.\ FW tolerance $\varepsilon$ at $N{=}20$ (left) and $N{=}50$ (right).
The x-axis is inverted so that the plot reads left-to-right as loose-to-exact; $\varepsilon{=}0$ (exact solution) is placed one decade beyond $10^{-8}$ for visibility.}
\label{fig:abl_epsilon}
\end{figure}

We sweep the FW stopping tolerance $\varepsilon\in\{0,10^{-8},10^{-5},10^{-2},10^{-1}\}$ for HullFT and for the Carath\'{e}odory-integerized variant.
\Cref{fig:abl_epsilon} reports BPB\% under this tolerance sweep at $N{=}20$ and $N{=}50$.
The x-axis is \emph{inverted}: the left end is the loosest tolerance ($\varepsilon{=}10^{-1}$) and the right end is the exact solution ($\varepsilon{=}0$), so moving right means strictly tighter convergence.
HullFT's BPB\% is flat across the entire range up through $\varepsilon{=}10^{-2}$, indicating that the support cap $m$, not the tolerance, is the binding constraint in practice. Meanwhile, the Carath\'{e}odory variant degrades noticeably once $\varepsilon{>}10^{-5}$, justifying the moderate tolerance used in the main experiments.

\subsection{FW variant: forced-unique selection without \texorpdfstring{$\varepsilon$}{epsilon} early-stop}
\label{sec:app_abl_fw_no_eps}
We ask whether the $\varepsilon$ early-stop and the allowance for point revisits in standard HullFT are important.
The \texttt{fw\_no\_epsilon} variant disables the $\varepsilon$ early-stop and forces each FW iteration to select a previously unchosen point.
When FW keeps adding distinct points, the weights assigned to them become spread thinly, so many of the selected points carry negligible weight and contribute little information.
Geometric integerization handles this gracefully: by selecting only the most important points from the support, it recovers performance on par with standard HullFT.
Variants that use the full point set without this selection step do not benefit from the additional points and perform noticeably worse.
\Cref{tab:abl_fw_no_epsilon} reports total runtime and BPB\% relative to the non-finetuned baseline at $N\in\{20,25,30,35\}$; all four variants are within $0.15s$ of each other.
\begin{table}[H]
\centering
\small
\caption{Total runtime (seconds) and BPB\% relative to the non-finetuned baseline (lower is better) for forced-unique Frank--Wolfe variants across different selection budgets $N$.}
\label{tab:abl_fw_no_epsilon}
\begin{tabular}{l c c c c}
\toprule
\textbf{Method} & \textbf{$N=20$} & \textbf{$N=25$} & \textbf{$N=30$} & \textbf{$N=35$} \\
\midrule
FW no-epsilon & $1.93s$ ($81.36\%$) & $2.27s$ ($80.80\%$) & $2.58s$ ($80.22\%$) & $2.88s$ ($79.98\%$) \\
FW no-epsilon + pad-by-weights & $1.98s$ ($81.60\%$) & $2.32s$ ($80.82\%$) & $2.63s$ ($80.38\%$) & $2.93s$ ($80.14\%$) \\
FW no-epsilon + integerize & $1.78s$ ($74.03\%$) & $2.13s$ ($71.93\%$) & $2.47s$ ($70.44\%$) & $2.81s$ ($69.44\%$) \\
HullFT (ours) & $1.81s$ ($74.04\%$) & $2.17s$ ($72.00\%$) & $2.52s$ ($70.52\%$) & $2.84s$ ($69.45\%$) \\
\bottomrule
\end{tabular}
\end{table}

\subsection{kNN pre-selection pool size \texorpdfstring{$k$}{k}}
\label{sec:app_abl_knn}
\begin{figure}[H]
\centering
\includegraphics[width=\linewidth]{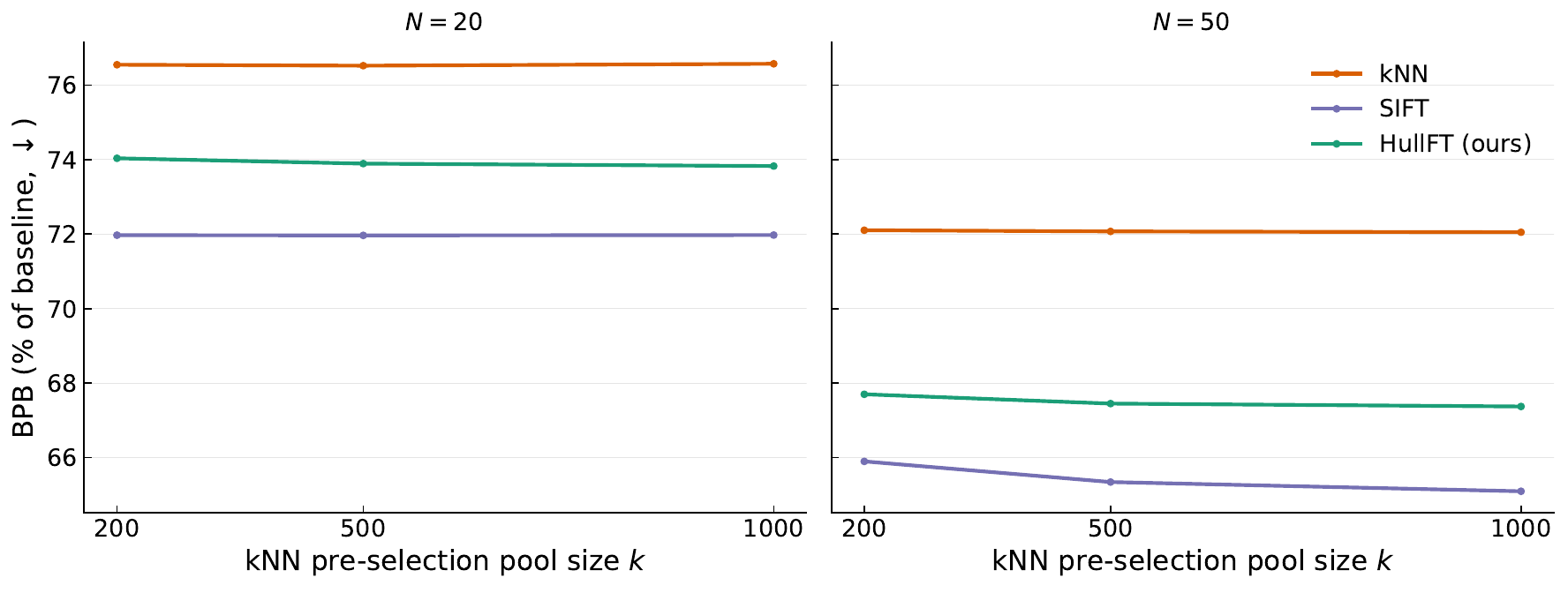}
\caption{Effect of the kNN pre-selection pool size $k$ on BPB\% at $N{=}20$ (left) and $N{=}50$ (right).}
\label{fig:abl_knn_preselect}
\end{figure}

We vary the pre-selection pool $k$ fed to the three selectors at two representative budgets $N\in\{20,50\}$.
\Cref{fig:abl_knn_preselect} reports BPB\% as the pre-selection pool size varies at each of these budgets.
All methods improve slightly with larger $k$, with returns saturating by $k{=}500$; HullFT stays strictly between kNN and SIFT at matched $N$ and remains Pareto-dominant on total runtime at every $k$.

\subsection{Per-subset breakdown against each baseline individually}
\label{sec:app_per_subset_vs_each}

\begin{figure}[H]
\centering
\includegraphics[width=\linewidth]{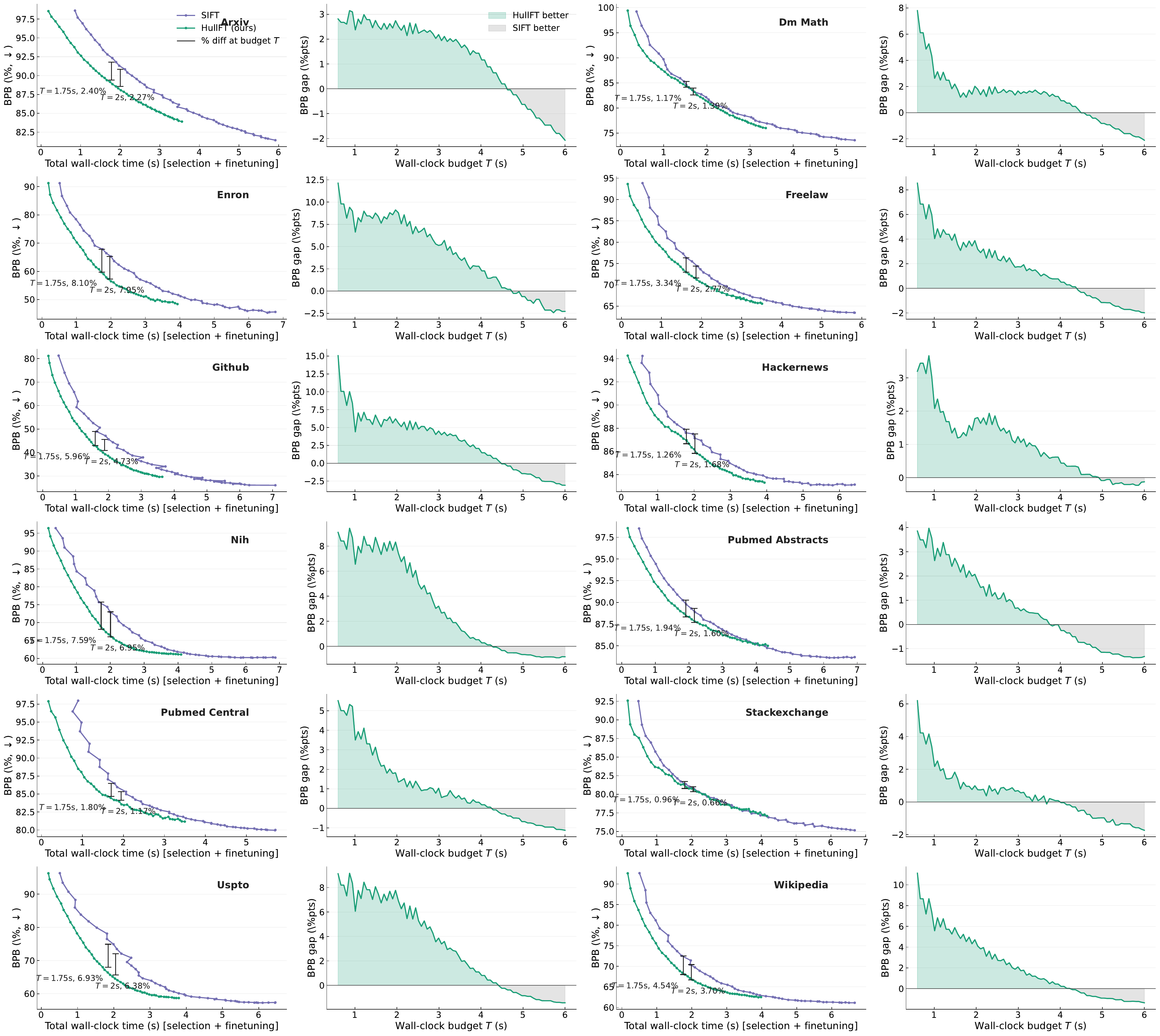}
\caption{Per-subset breakdown against \textbf{SIFT} for all 12 Pile subsets (2 subsets per row).
For each subset we show, \emph{left}: BPB\% vs.\ wall-clock time, and \emph{right}: the quality gap (SIFT $-$ HullFT) as a function of the total-runtime budget $T$.}
\label{fig:per_subset_all_vs_sift}
\end{figure}

\begin{figure}[H]
\centering
\includegraphics[width=\linewidth]{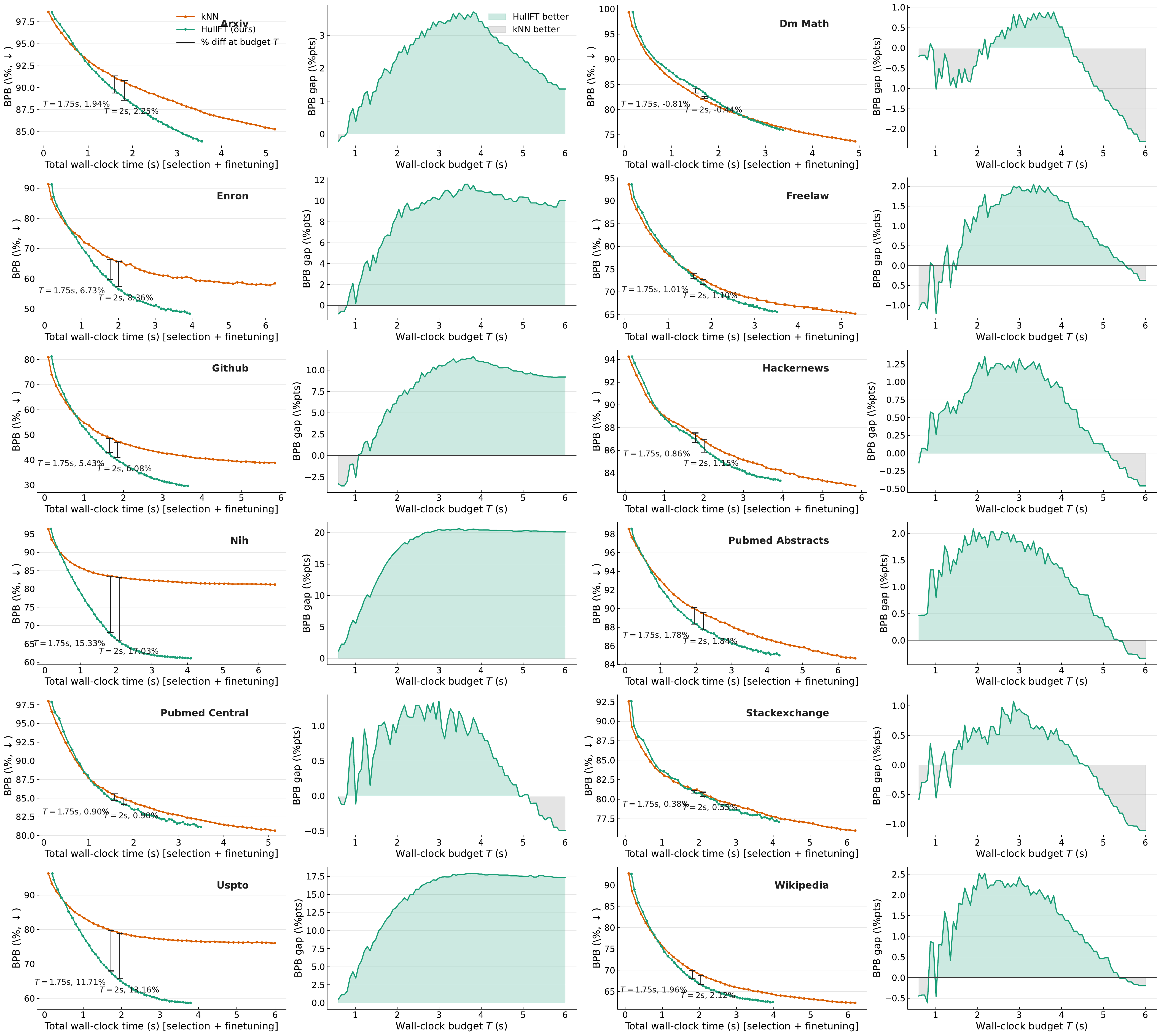}
\caption{Per-subset breakdown against \textbf{kNN} for all 12 Pile subsets (2 subsets per row).
For each subset we show, \emph{left}: BPB\% vs.\ wall-clock time, and \emph{right}: the quality gap (kNN $-$ HullFT) as a function of the total-runtime budget $T$.}
\label{fig:per_subset_all_vs_knn}
\end{figure}

\Cref{fig:per_subset_all} in the main text reports the per-subset gap against the \emph{best} of kNN and SIFT at each budget.
For completeness, \cref{fig:per_subset_all_vs_sift,fig:per_subset_all_vs_knn} show the same breakdown against each baseline in isolation, so the reader can see where HullFT's advantage comes from kNN vs.\ SIFT separately.
At each anchor $T$, $N^{\star}$ is taken from the global aggregate (matching \cref{tab:per_subset_budgets}) and per-subset BPB\% is read at those fixed $N^{\star}$ values.

\subsection{Fidelity of geometric integerization to Frank--Wolfe weights}
\label{sec:app_integerization_distance}

\begin{figure}[H]
\centering
\includegraphics[width=\linewidth]{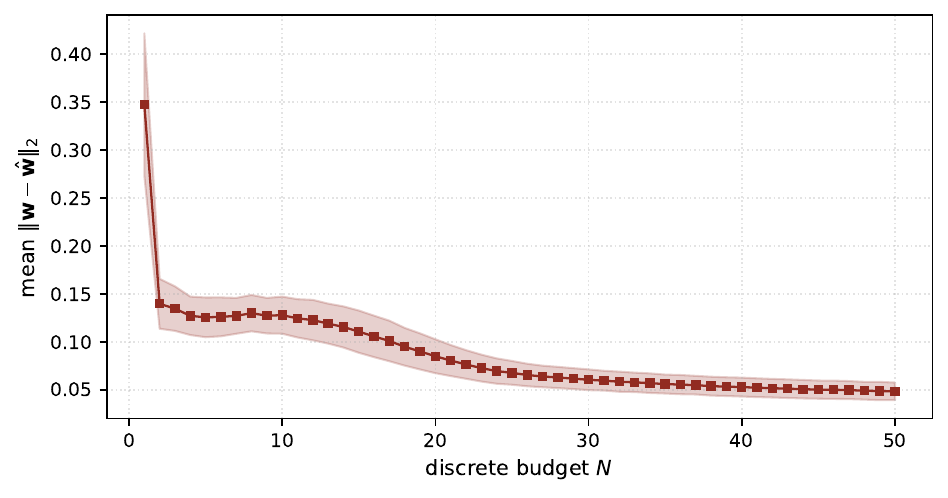}
\caption{Fidelity of geometric integerization to Frank--Wolfe weights.
Mean $\ell_2$ distance between FW weights $\mathbf{w}$ on the support and empirical weights $\hat{\mathbf{w}}$ with $\hat{w}_i=c_i/N$.
Each point averages over test queries within a subset, then over twelve Pile subsets; the band shows $\pm$ one standard deviation across subsets.}
\label{fig:integerization_distance}
\end{figure}

Frank--Wolfe produces fractional weights on a sparse support; geometric integerization converts these into an exact $N$-tuple of pool indices with multiplicities $c_i$, so the induced empirical weights are $\hat{w}_i = c_i/N$ on that same support.
We quantify how much the discrete allocation departs from the pre-integerization FW vector by the per-query distance $\|\mathbf{w}-\hat{\mathbf{w}}\|_2$, averaged over all test queries in each subset.
\cref{fig:integerization_distance} reports the mean of these averages across the twelve Pile subsets for which we precomputed embeddings (shaded band: $\pm$ one standard deviation across subsets).
The distance stays in a moderate range for every $N\le 50$, which supports the view that integerization preserves the FW weight profile rather than inducing an unrelated discrete selection.

\subsection{Selection geometry: t-SNE visualizations}
\label{sec:app_tsne_distance}

\begin{figure}[H]
\centering
\includegraphics[width=\linewidth]{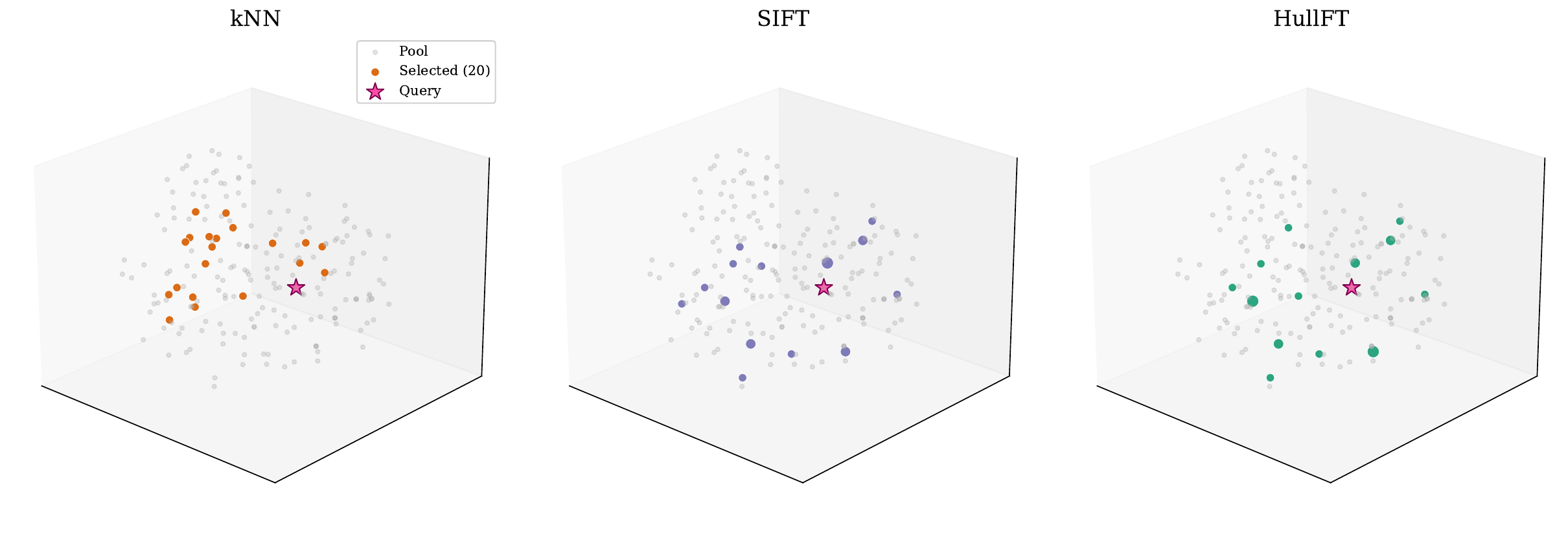}
\vspace{0.5em}
\includegraphics[width=\linewidth]{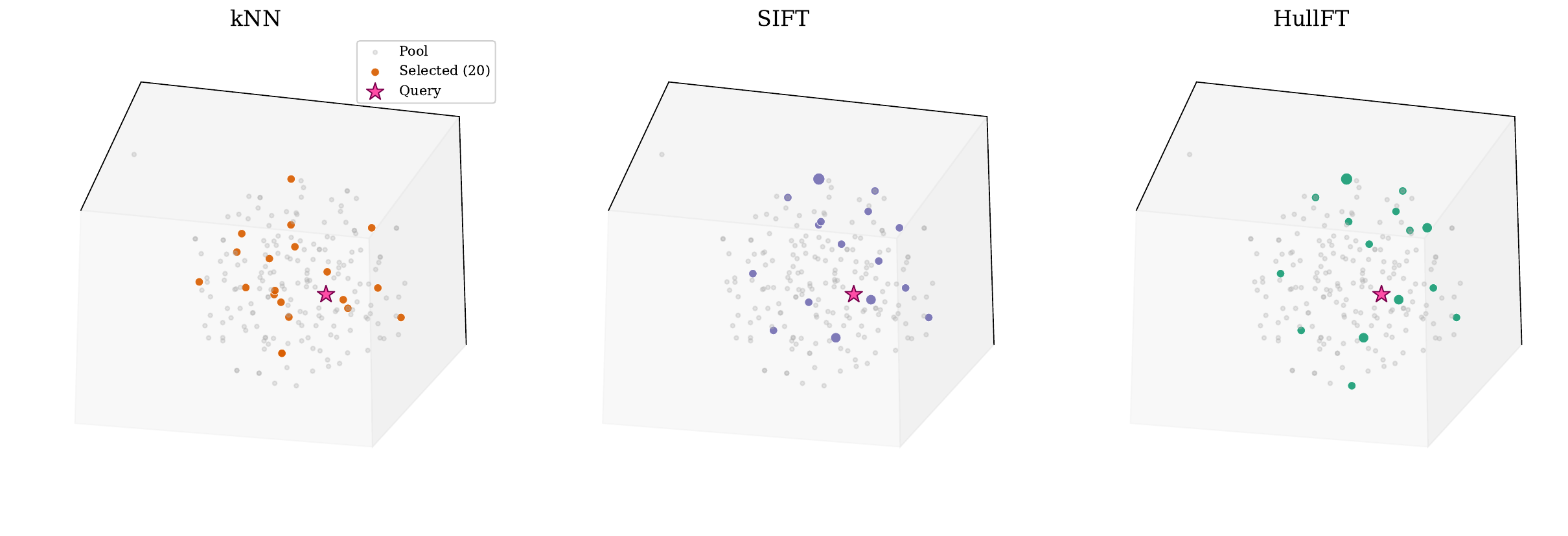}
\caption{3D t-SNE projections of the candidate pool (gray) and the $N{=}20$ points selected by each method (colored) for a representative test query (pink star) from GitHub (top) and FreeLaw (bottom). HullFT delivers a selection geometry comparable to SIFT through convex reconstruction, at substantially lower selection latency (\cref{sec:cpu_test}).}
\label{fig:tsne_comparison}
\end{figure}

We extend the qualitative analysis of the main text to two additional subsets, GitHub and FreeLaw, which exhibit different candidate-pool structure.
\Cref{fig:tsne_comparison} shows a 3D t-SNE projection of the query, candidate pool, and the $N{=}20$ points returned by kNN, SIFT, and HullFT on a representative query from each subset.
The pattern observed for ArXiv recurs: kNN clusters tightly around the query, SIFT places its points more broadly around the query to reduce redundancy, and HullFT produces a comparably diverse support obtained directly from the convex approximation in \cref{sec:method}.
Because the same behavior appears across pools with markedly different topology, the diversity is most plausibly attributable to the geometric mechanism itself rather than to properties specific to a single domain.
This diversity is achieved at a fraction of SIFT's per-query selection cost (\cref{sec:cpu_test}), reinforcing the headline efficiency--quality tradeoff reported in the main text.

\subsection{Qualitative examples of selected contexts}
\label{sec:app_qualitative}

For two representative test queries in each of ArXiv, GitHub, and PubMed Central, we show the test query alongside the two candidates with the largest weight in the HullFT support at $N{=}20$.
All text below is verbatim from the Pile candidate pool; we truncate each excerpt to roughly 400 characters for readability.

\paragraph{ArXiv.}\mbox{}\par
\begin{center}
\footnotesize
\begin{tabular}{p{0.16\linewidth} p{0.78\linewidth}}
\toprule
\textbf{prompt} & The operational meaning of quantum conditional information. Information might be regarded as an answer to a question. To know if it is raining, one need only look outside. However, a person living in a desert would expect a different answer than one living in a climate where on average, it rains every other day. After looking outside, who gains ``more'' information? The answer to this question has nothing to do with weather $\dots$ \\
\midrule
Example 1 & Every forecasting system is noncalibrated for uncountably many data sequences that it might see. This result is strengthened here: from a topological point of view, failure of calibration is typical and calibration rare. Meanwhile, Bayesian forecasters are certain that they are calibrated---this invites worries about the connection between Bayesianism and rationality. $\dots$ \\
\midrule
Example 2 & Landauer's erasure principle exposes an intrinsic relation between thermodynamics and information theory: the erasure of information stored in a system, $S$, requires an amount of work proportional to the entropy of that system. This entropy, $H(S|O)$, depends on the information that a given observer, $O$, has about $S$ $\dots$ \\
\midrule
\textbf{prompt} & Nanoscale particles embedded in a metallic matrix are of considerable interest as a route towards identifying and tailoring material properties. We present a detailed investigation of the electronic structure, and in particular the Fermi surface, of a nanoscale phase ($L1_2$ Al$_3$Li) that has so far been inaccessible with conventional techniques $\dots$ \\
\midrule
Example 1 & Point defect species and concentrations in metastable Fe--C alloys are determined using density functional theory and a constrained free-energy functional. Carbon interstitials dominate unless iron vacancies are in significant excess, whereas excess carbon causes greatly enhances vacancy concentration $\dots$ \\
\midrule
Example 2 & The Fermi surface topology of the shape-memory alloy Ni$_{0.62}$Al$_{0.38}$ has been determined using Compton scattering. A large area of this Fermi surface can be made to nest with other areas by translation through a vector of $\approx 0.18\,[1,1,0](2\pi/a)$, which corresponds to the wavevector associated with martensitic precursor phenomena $\dots$ \\
\bottomrule
\end{tabular}
\end{center}

\paragraph{GitHub.}\mbox{}\par
\begin{center}
\footnotesize
\begin{tabular}{p{0.16\linewidth} p{0.78\linewidth}}
\toprule
\textbf{prompt} & \texttt{// Debugger.Object referents can be transparent wrappers of objects in the debugger compartment. var g = newGlobal(); g.f = function (a, b) \{ return a + "/" + b; \}; var dbg = Debugger(g); var hits = 0; dbg.onDebuggerStatement = function (frame) \{ var f = frame.eval("f").return; assertEq(f.call(null, "a", "b").return, "a/b"); hits++; \}; g.eval("debugger;"); assertEq(hits, 1);} \\
\midrule
Example 1 & \texttt{// setVariable can set variables and arguments in functions. var g = newGlobal(); var dbg = new Debugger(g); dbg.onDebuggerStatement = function (frame) \{ frame.environment.setVariable("a", 100); frame.environment.setVariable("b", 200); \}; g.eval("function f(a) \{ var b = a + 1; debugger; return a + b; \}"); assertEq(g.f(1), 300);} \\
\midrule
Example 2 & \texttt{// Referents of Debugger.Objects in other compartments always survive per-compartment GC. var g = newGlobal(); var dbg = Debugger(g); var arr = []; dbg.onDebuggerStatement = function (frame) \{ arr.push(frame.eval("[]").return); \}; g.eval("for (var i = 0; i < 10; i++) debugger;"); assertEq(arr.length, 10); gc(g);} \\
\midrule
\textbf{prompt} & \texttt{class Foo \{ // function fBar (x,y); fOne(argA, argB, argC, argD, argE, argF, argG, argH) \{ Array<string> numbers = ['one','two','three','four','five','six']; var x = ("" + argA) + argB + argC + argD + argE + argF + argG + argH; try \{ this.fTwo(argA, argB, argC, this.fThree("", argE, argF, argG, argH)); \} catch (e,s) \{\} $\dots$} \\
\midrule
Example 1 & \texttt{class Foo \{ // function fBar (x,y); fOne(argA, argB, argC, argD, argE, argF, argG, argH) \{ Array<string> numbers = ['one','two','three','four','five','six']; var x = ("" + argA) + argB + argC + argD + argE + argF + argG + argH; try \{ this.fTwo(argA, argB, argC, this.fThree("", argE, argF, argG, argH)); \} catch (error, stack) \{\} $\dots$} \\
\midrule
Example 2 & \texttt{import com.Foo; using com.utils.MathUtil; class Foo \{ // function fBar (x,y); function fOne(argA:Int, argB, argC, argD, argE, argF, argG, argH) \{ var numbers:Array<String> = ['one','two','three','four','five','six']; var x = ("" + argA) + argB + argC + argD + argE + argF + argG + argH; $\dots$} \\
\bottomrule
\end{tabular}
\end{center}

\paragraph{PubMed Central.}\mbox{}\par
\begin{center}
\footnotesize
\begin{tabular}{p{0.16\linewidth} p{0.78\linewidth}}
\toprule
\textbf{prompt} & Childhood psychiatric disorders are of concern because of the associated negative impact on general wellbeing across the lifespan, for example, lower educational achievements. The findings in several studies suggest that the adverse impact of childhood onset psychiatric disorders on educational attainment is largely accounted for by problems of inattention and conduct $\dots$ \\
\midrule
Example 1 & The costs of primary and lower secondary schooling in Denmark are among the highest in OECD. This is in part due to large amounts being spent on children in special needs education; about one third of all the costs of primary and lower secondary education $\dots$ \\
\midrule
Example 2 & Juvenile idiopathic arthritis (JIA) is an umbrella term for chronic childhood arthritis and a significant contributor to chronic disease in children and adolescents. Even in the era of modern treatment with improved outcome on biologic therapies, many children with JIA still experience flares and difficulties to attend daily life activities $\dots$ \\
\midrule
\textbf{prompt} & Dengue fever (DF) is an insect-borne disease caused by four different dengue viruses (DENV 1--4), which are mainly transmitted by \emph{Ae.~aegypti} and \emph{Ae.~Albopictus}. DF is endemic to more than 100 countries in tropical and subtropical areas, especially in Southeast Asia, the Americas, the Western Pacific, Africa $\dots$ \\
\midrule
Example 1 & Dengue is a vector-borne tropical disease caused by the four dengue virus serotypes (DENV 1--4) that is mainly transmitted by \emph{Aedes aegypti} and \emph{Aedes albopictus} mosquitoes. Dengue is an endemic infectious disease in most of the tropical and subtropical regions, especially in Southeast Asia, the Western Pacific, Latin America $\dots$ \\
\midrule
Example 2 & Dengue fever, one of the most prevalent mosquito-borne diseases in humans, is mainly transmitted by \emph{Aedes aegypti} and \emph{Aedes albopictus}. There are four distinct serotypes for dengue virus, namely DENV 1, 2, 3 and 4. Dengue fever is endemic in more than 100 countries of Southeast Asia $\dots$ \\
\bottomrule
\end{tabular}
\end{center}


\end{document}